\documentclass[final,3p,times,twocolumn,authoryear]{elsarticle}

\usepackage{amssymb}
\usepackage{amsmath,amsfonts}
\usepackage{booktabs}
\usepackage{subfigure}
\usepackage{amsthm}
\usepackage{multicol}
\usepackage{multirow}
\usepackage{algorithm}
\usepackage{algpseudocode}

\newtheorem{theorem}{Theorem}

\usepackage{color}

\newcommand{\nosection}[1]{\vspace{2pt}\noindent\textbf{#1.}}

\newcommand{\lyy}[1]{\textcolor{black}{#1}}

\newcommand{\model}{FedAF}

\usepackage{apalike}
\usepackage{url}

\journal{Knowledge-Based Systems}

\begin{document}

\begin{frontmatter}



\title{Class-wise Federated Unlearning: Harnessing Active Forgetting with Teacher-Student Memory Generation}

\author[hduce]{Yuyuan Li}
\ead{y2li@hdu.edu.cn}

\author[zjucs]{Jiaming Zhang}
\ead{22321350@zju.edu.cn}

\author[hducy]{Yixiu Liu}
\ead{liuyixiu@hdu.edu.cn}

\author[zjucs]{Chaochao Chen\corref{cor1}}
\ead{zjuccc@zju.edu.cn}
\cortext[cor1]{Corresponding author}

\address[hduce]{School of Communication Engineering, Hangzhou Dianzi University, Hangzhou 310018, China}
\address[zjucs]{College of Computer Science \& Technology, Zhejiang University, Hangzhou 310027, China}
\address[hducy]{School of Cyberspace, Hangzhou Dianzi University, Hangzhou 310018, China}

\begin{abstract}
Privacy concerns associated with machine learning models have driven research into machine unlearning, which aims to erase the memory of specific target training data from already trained models. This issue also arises in federated learning, creating the need to address the federated unlearning problem. However, federated unlearning remains a challenging task. On the one hand, current research primarily focuses on unlearning all data from a client, overlooking more fine-grained unlearning targets, e.g., class-wise and sample-wise removal. On the other hand, existing methods suffer from imprecise estimation of data influence and impose significant computational or storage burden. To address these issues, we propose a neuro-inspired federated unlearning framework based on active forgetting, which is independent of model architectures and suitable for fine-grained unlearning targets. Our framework distinguishes itself from existing methods by utilizing new memories to overwrite old ones. These new memories are generated through teacher-student learning. We further utilize refined elastic weight consolidation to mitigate catastrophic forgetting of non-target data. Extensive experiments on benchmark datasets demonstrate the efficiency and effectiveness of our method, achieving satisfactory unlearning completeness against backdoor attacks.
\end{abstract}


\end{frontmatter}


\section{Introduction}

The release of data regulations, e.g., General Data Protection Regulation~\citep{euro2018gdpr}, Califonia Consumer Privacy Act~\citep{cal2018del}, and Delete Act~\citep{cal2023del}, has significantly influenced privacy standards and data management practices in the field of machine learning. 
The growing accumulation of personal data has heightened privacy concerns, prompting the adoption of decentralized learning frameworks like federated learning~\citep{mcmahan2017communication}, which enables model training on distributed data without centralizing sensitive information.
In addition to privacy concerns, the \textit{Right To Be Forgotten} (RTBF) has garnered substantial academic attention, as it grants individuals the authority to remove their personal data from systems.
This right has fundamentally reshaped data management paradigms.
Given the potential of machine learning models to memorize training data~\citep{bourtoule2021machine, li2023making}, complying with the RTBF necessitates not only the deletion of data from databases but also the elimination of any residual memory of the data within models.
This requirement has spurred the development of machine unlearning techniques, which aim to remove the influence of specific data samples from trained models~\citep{nguyen2022survey}.
The intersection of privacy concerns and the RTBF has given rise to federated unlearning, which focuses on achieving machine unlearning within the context of federated learning scenarios~\citep{liu2023survey}.

\begin{figure*}[t]
    \centering
    \includegraphics[width=\linewidth]{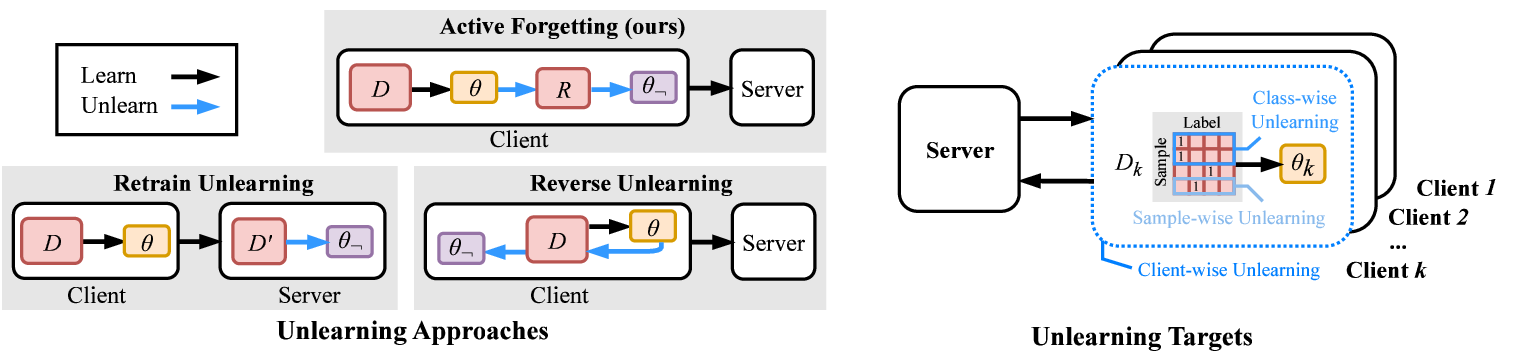}
    \caption{An overview of different federated unlearning approaches (right) and targets (left). $\theta$ and $\theta_{\lnot}$ denote original and unlearned models respectively. $\mathcal{D}$, $\mathcal{R}$, and $\mathcal{D}'$ denote the original training dataset, the target data (to be unlearned), and the updated dataset without the target data ($\mathcal{D} \backslash \mathcal{R}$) respectively.}
    \label{fig:comparison}
\end{figure*}

However, existing federated unlearning methods suffer from certain limitations. 
Firstly, the majority of these methods primarily focus on client-wise unlearning, which removes all data of a client (as shown in the right of  Figure~\ref{fig:comparison})~\citep{gao21federaser,liu2022right,wei23federated}. This focus neglects more fine-grained unlearning targets, e.g., class-wise and sample-wise removal. 
This limitation constrains the ability to precisely remove specific patterns or instances from trained models.
Moreover, methods devised for fine-grained targets can scale to client-wise unlearning, but the reverse is not true.
On the other hand, client-wise unlearning methods also face challenges, including imprecise approximation~\citep{basu2021influence} and storage burden~\citep{guo2021fastif}.
Secondly, existing limited fine-grained unlearning methods face challenges such as imprecise estimation of data influence~\citep{wu2022federated,che2023fast} or restricted applicability for specific models, e.g., CNN~\citep{wang2022federated,xia2023fedme} and knowledge graph~\citep{zhu2023heterogeneous}. 
These challenges hinder their accuracy and scalability.

In this paper, we propose a Federated Active Forgetting (\model{}) framework, which is scalable for fine-grained targets, and independent of specific model architectures.
As shown in Figure~\ref{fig:comparison}, \model{} presents a distinct design compared to the prevailing unlearning approaches.
Drawing inspiration from \textit{active forgetting} in neurology~\citep{davis2017biology,ryan2022forgetting}, \model{} allows the target client, i.e., the client intending to unlearn, to use new memories to overwrite specific old ones. This process mirrors the function of dopamine-producing forgetting cells that accelerate the elimination of memorization.
This design facilitates a smooth unlearning process that can be seamlessly integrated into the original learning procedure, which fundamentally overcomes the limitations of existing federated unlearning methods. Note that it eliminates the need for storing historical updates or estimating data influence.

We implement the design of active forgetting based on incremental learning, which focuses on the task of sequentially learning multiple tasks~\citep{wang2021afec}.
Specifically, the original learning process and the subsequent unlearning process can be regarded as two sequential tasks.
For simplicity, we treat all unlearning requests as a single process.
The key distinction between incremental learning and federated unlearning lies in the task conflict. 
In incremental learning, new knowledge is acquired without conflict between two sequential tasks
In contrast, federated unlearning involves a conflict, as the subsequent task requires unlearning the previously acquired knowledge from the previous task.
To address this conflict, we derive a new loss function for model training.

When utilizing incremental learning for unlearning, two critical considerations arise: (i) generating effective new memories, and (ii) overcoming catastrophic forgetting.
Our proposed \model{} addresses these issues through two modules: the \textit{memory generator} and the \textit{knowledge preserver}.
First, we aim to generate knowledge-free and easy-to-learn new memories to overwrite the old ones.
``Knowledge-free'' means the new memories do not contain meaningful information about the data, e.g., a random label vector.
Meanwhile, these new memories have to be easy-to-learn so that it will not cost considerable computational overhead to accomplish unlearning.
The memory generator is designed to follow a teacher-student learning paradigm to generate fake labels, which are then paired with the original features.
Second, previous studies found that conventional deep learning methods struggle with incremental learning due to the phenomenon of catastrophic forgetting~\citep{chen2021overcoming}.
This phenomenon involves the unintended removal of both target and non-target data memories.
The knowledge preserver alleviates the issue of unintentional removal by elastically constraining the model's parameters with our derived loss. 
The main contributions of this paper are summarized as follows:
\begin{itemize}
    \item We propose a neuro-inspired federated unlearning method (\model{}) based on the concept of active forgetting, which is scalable for class-wise unlearning and independent of specific model architectures. 
    \item To effectively generate new memories to overwrite the old ones, we adopt a teacher-student learning paradigm.
    The target client, i.e., the student, distills knowledge from manipulated data that teacher models generate.
    \item To alleviate the catastrophic forgetting of non-target data, we first derive a new loss to address the problem of task conflict, and then dynamically constrain the change of model's parameters. 
    \item We conduct extensive experiments across three benchmark datasets to evaluate the performance of \model{}.
    The results show that our proposed \model{} generally outperforms compared methods in terms of efficiency, utility, and completeness.
\end{itemize}

\section{Preliminaries}

In this section, we first introduce the notations of federated learning and unlearning, followed by the principles of federated unlearning. 
Afterward, we clarify the unlearning target in this paper.

\subsection{Notation}\label{sec:nota}

\nosection{Federated Learning}\label{sec:fl} Federated learning allows multiple clients to collaboratively train a global model without sharing their private data. 
A typical architecture of federated learning, i.e., FedAvg~\citep{mcmahan2017communication} can be formulated as follows:
There are $K$ clients and a server.
Each client, denoted as $k \in [K]$, possesses a local dataset $\mathcal{D}_k$ containing $n_k$ samples.
The goal is to obtain a global model parameterized by $\theta^*$ that minimizes the empirical risk over all clients:
\begin{equation}
    \theta^* = \arg\min_{\theta} \frac{1}{K} \sum_{k=1}^{K} w_k L_k(\theta),
\end{equation}
where $w_k = n_k/\sum_k n_k$ is the weight assigned to client $k$, and $L_k(\theta)$ is the local loss function of client $k$.
At $t$-th federated iteration, each client $k$ performs $E$ epochs of stochastic gradient descent on its local dataset $\mathcal{D}_k$ using the current global model parameters $\theta_t$, and then uploads the updated model parameters $\theta_{k, t+1}$ to the server for aggregation.
The server aggregates the model updates from all clients using a weighted average:
\begin{equation}\label{equ:agg}
    \theta_{t+1} = \sum_{k=1}^K = w_k \theta_{k, t+1}.
\end{equation}
This process continues until the global model $\theta$ converges or reaches the maximum federated iterations.

\nosection{Federated Unlearning} 
As formulated above, each client $k$ possesses a local dataset $\mathcal{D}_k$.
Each client has the RTBF to remove any of its own data and corresponding influence in the global model.
This process is termed as unlearning.
Formally, the client $k$ submits a request to unlearn a specific target $\mathcal{R} \subseteq \mathcal{D}_k$.
Following~\citep{liu2022right,gao2022verifi}, we assume that the requests are submitted after the end of federated training, i.e., the global model has reached convergence.
In practice, clients may submit unlearning requests during federated training, but evaluating the model performance before and after unlearning during federated training is intractable.
After receiving the requests, the server instructs the clients to unlearn the target data while keeping their local data private. 
We denote the parameters of the ground-truth unlearned model by $\theta_{\lnot}^*$ which is collaboratively retrained across all clients on $\mathcal{D} \backslash \mathcal{R}$ from scratch. 

\subsection{Unlearning Principles}
\label{sec:pri}
We identify four principles that we consider as the pillars of achieving successful federated unlearning.
Similar objectives of the first three principles can also be found in previous studies~\citep{bourtoule2021machine,li2024making}.

\begin{itemize}
    \item \textbf{P1: Unlearning Completeness.} Complete unlearning refers to entirely removing the influence, i.e., memory, of target data, making it irretrievable.
In some real-world scenarios, completeness may be partially sacrificed for efficiency.
    \item \textbf{P2: Unlearning Efficiency.} Practical ML models often involve large-scale datasets and parameters, which results in significant computational overheads in terms of both time and space.
As a consequence, retraining from scratch is prohibitive and efficiency is an important principle.
    \item \textbf{P3: Model Utility.} It is evident that both clients and servers desire to maintain model performance after unlearning.
However, it is important to acknowledge that unlearning a significant amount of data lineage would inevitably diminish the model utility, because unlearning is equivalent to reducing the amount of training data. 
Thus, an adequate unlearning method needs to generate an unlearned model that achieves comparable performance to a model retrained from scratch $\theta_{\lnot}^*$. 
    \item \textbf{P4: Data Privacy.} Due to the inherent requirement of federated learning, it becomes imperative to ensure that user privacy, i.e., sharing raw data, is not compromised during federated unlearning. 
\end{itemize}

\subsection{Unlearning Targets} 

As shown in Figure~\ref{fig:comparison} (right), unlearning targets can be mainly classified into four categories based on their scope, i.e., client-wise (all data of a client), class-wise (all data of a class), sample-wise (a data sample), and feature-wise (one feature dimension of a data sample).
In federated learning, clients possess data without permission to share.
Thus, in this paper, class-wise target refers to a specific class of data that is possessed by the target client.
The capabilities of unlearning targets with a smaller scope can be transferred to a larger scope, but the opposite may not be true.
Our proposed \model{} can handle all three types of targets and is particularly suitable for class-wise targets.
To fully exploit the capability of \model{}, in this paper, we mainly conduct experiments of class-wise unlearning, without compromising the generality.

\section{Related Work}

In this section, we provide a brief review of the existing literature on machine unlearning and federated unlearning, followed by a summary of incremental learning methods.

\subsection{Machine Unlearning}

Based on the degree of unlearning completeness, machine unlearning methods can be categorized into two approaches: exact unlearning (full completeness) and approximate unlearning (partial completeness). 
The choice of approach depends on the situation and desired outcome.

\nosection{Exact Unlearning (EU)} This approach aims to completely remove the memory of target data.
Deleting the target data and retraining the model on the remaining data from scratch is a naive but algorithmic unlearning approach, i.e., naturally enabling full completeness.
As retraining from scratch takes considerable computational overhead, EU approach mainly focuses on enhancing retraining efficiency.
The main idea behind existing EU methods is partition-aggregation~\citep{ginart2019making,schelter2021hedgecut,bourtoule2021machine,yan2022arcane,li2023ultrare}, which involves partitioning the dataset or model into sub-components, training them individually, and then aggregating them at the end.
With this idea, EU methods limit the overhead of retraining to sub-components.

\nosection{Approximated Unlearning (AU)} This approach aims to expedite unlearning by approximating the parameters of the ground truth model, i.e., retraining from scratch.
Existing AU methods estimate the influence of target data and directly remove it through reverse gradient operations~\citep{sekhari2021remember,wu2022puma,mehta2022deep, li2023selective}. 
However, precise estimation of influence and managing the computational overhead remain an ongoing challenge for deep learning models~\citep{basu2021influence,guo2021fastif}.

\subsection{Federated Unlearning}

Existing federated unlearning methods predominantly focus on AU, because implementing EU, i.e., partition-aggregation frameworks, in a distributed manner does not yield substantial improvements in time efficiency.
Note that in the context of federated learning, the data is already distributed across clients.
Federated unlearning methods can be further divided into the following three categories, i.e., retrain unlearning, reverse unlearning, and others.

\nosection{Retrain Unlearning} This approach mimics EU approach to achieve efficient retraining.
It accelerates retraining by approximating the gradients using previously stored historical updates~\citep{gao21federaser,liu2022right,wei23federated}.
This approach is hindered by inaccurate approximation and the burden of data storage.

\nosection{Reverse Unlearning} This approach follows the idea of AU, removing the estimated influence through reverse gradient operations, e.g., loss maximization~\citep{anisa2022federated}, gradient ascent~\citep{wu2022federated}, and Nemytskii operator~\citep{che2023fast}.
Analogous challenges to the AU approach are encountered by this approach.

\nosection{Others} There are other model-specific federated unlearning methods focusing on CNN~\citep{wang2022federated,xia2023fedme}, knowledge graph~\citep{zhu2023heterogeneous}, and Bayesian neural networks~\citep{wang2023bfu}.

As for comparison, our proposed \model{} is based on active forgetting that continues the original federated learning process to achieve unlearning.
This unlearning scheme not only establishes a seamless learning--unlearning workflow, but is also independent of model architectures.

\subsection{Incremental Learning}

Incremental learning, also known as continual learning
and lifelong learning in literature, aims to sequentially learn multiple tasks~\citep{li2019learn}.
Note that (i) transfer learning focuses only on the performance of the target task, which refers to only the unlearning data in our setting, and (ii) Multi-task learning focuses on the performance of all tasks, but it requires the data of all tasks at the same time, which is against our setting.
Different from transfer learning and multi-task learning, incremental learning focuses on achieving high performance across all tasks while only having access to the data of new task(s).
This aligns with the setting of federated unlearning, where it can be arduous to reach for historical updates.
There are mainly three approaches to address catastrophic forgetting in incremental learning:
\begin{itemize}
    \item \textit{Selective Synaptic Plasticity} elastically constrains parameter change based on synaptic importance to preserve learned knowledge~\citep{aljundi2018memory,kolouri2020sliced}.
    It is known for Elastic Weight Consolidation (EWC) framework.
    \item \textit{Additional Neural Resource Allocation} allocates new parameters for new tasks~\citep{schwarz2018progress,li2019learn}.
    This approach changes the model structure, which is obviously unsuitable for the unlearning problem.
    \item \textit{Memory Replay} stores previous data or generates pseudo-data, and replays this data with the new data~\citep{wu2018memory,hu2019overcoming}.
    This approach either increases extra computational overhead or changes the original learning process.
    Thus it is out of our consideration.
\end{itemize}

\section{Active Forgetting Framework}

\begin{figure*}[t]
    \centering
    \includegraphics[width=\linewidth]{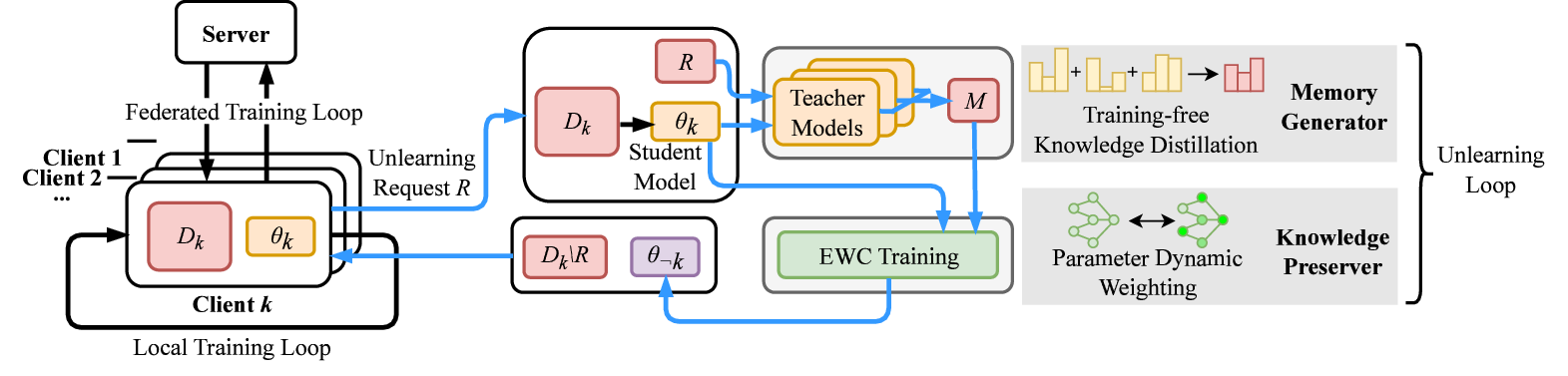}
    \caption{An overview of \model{} which integrates the unlearning loop into the local training loop of a client.
    The unlearning loop consists of a memory generator and a knowledge preserver, where $\mathcal{M}$ denotes new memory (manipulated data), $\mathcal{D}$ is the dataset, $\mathcal{R}$ denotes unlearning request (target data), and $\theta$ denotes model parameters.}
    \label{fig:framework}
\end{figure*}


\nosection{Motivation} 
The challenge of federated unlearning arises due to the extensive collaboration between clients and the server in federated learning.
Existing federated unlearning methods suffer from imprecise data influence estimation and high computational burden. 
Inspired by \textit{active forgetting} in neurology, we propose a novel federated unlearning framework named \model{}.
Different from natural forgetting, i.e., passive forgetting, active forgetting can induce the forgetting of specific memories~\citep{anderson2021active}.

\nosection{Advantages} Forgetting can be interpreted as a form of adaptive learning due to neuroplasticity~\citep{ryan2022forgetting}.
To actively forget the data memory, our proposed \model{} framework continually trains the model with new memories, i.e., manipulated data, which are generated from the old memories that need to be forgotten, i.e., target data.
Compared with existing approaches, \model{} has the following advantages:
\begin{itemize}
    \item \model{}'s unlearning process can be considered as a part of the extended learning process, reducing the additional impact on the model utility to a minimal level.
    \item \model{} has wide scalability. 
    As it seamlessly integrates the unlearning process into the learning process, it is independent of specific model architectures and applicable for fine-grained unlearning targets.
    
    \item Compared with retrain unlearning, \model{} neither requires additional storage for historical updates nor spends extra computational overhead for non-target data, reducing the deployment cost of unlearning in practice.
    
    \item Compared with reverse unlearning, \model{} avoids estimating the influence of data, which is found analytically intractable in deep learning~\citep{graves2021amnesiac}, and continues the original training to achieve unlearning.
    This enables \model{} to unlearn with higher precision.
\end{itemize}

\subsection{Framework Overview}
\label{sec:over}
\model{} learns from the new memories to achieve unlearning, which overcomes the disadvantages of both retrain unlearning and reverse unlearning.
The new memories contain no effective knowledge of the target data, which can overwrite the old ones.
Figure~\ref{fig:framework} shows the learning (black arrows) and unlearning (blue arrows) workflow of \model{}.
In the learning workflow, \model{} neither interferes with the original federated learning process nor stores any historical update.
As described in Section~\ref{sec:fl}, we can condense the original federated learning process into two loops, i.e., local training loop and federated training loop.
In the unlearning workflow, \model{} incorporates an unlearning loop within the local training loop, and then broadcasts the unlearned update through the federated training loop.

The unlearning loop in \model{} consists of two modules, i.e., \textit{memory generator} and \textit{knowledge preserver}.
These two modules tackle the aforementioned issues respectively, i.e., generating effective new memories and overcoming catastrophic forgetting.
In general, the memory generator initializes a set of teacher models to generate fake labels, and then pairs them with the original feature to produce the manipulated data.
The knowledge preserver continually trains the model on the manipulated data.
With the help of the new loss function derived from EWC framework, the knowledge preserver manages to alleviate the catastrophic forgetting phenomenon.
Algorithm~\ref{alg:aff} summarizes the details in the unlearning workflow, where lines 1 to 6 represent the memory generator and line 7 represents the knowledge preserver.

\begin{algorithm}[tb]
\caption{Federated Active Forgetting Framework}
\label{alg:aff}
\textbf{Require}: Client $k$ submits an unlearning request $\mathcal{R} \subseteq \mathcal{D}_k$.\\
\textbf{Procedure}:
\begin{algorithmic}[1] 
\State Let new memories $\mathcal{M} = \emptyset$;
\State Initialize $Q$ teacher models without training;
\For{$(x, y$) in $\mathcal{R}$}
    \State Generate fake label $\tilde{y}$ from teacher models by Eq~(\ref{equ:debias});
    \State Add ($x, \tilde{y}$) to $\mathcal{M}$; \Comment{Teacher models can be released after use.}
\EndFor
\State Train $\theta$ on $\mathcal{M}$ and $\mathcal{R}$ by Eq~(\ref{equ:new}), produce the local unlearned model $\theta_{\lnot k}$;
\State Aggregate local unlearned model 
$\theta_{\lnot k}$ to the server by Eq~(\ref{equ:agg});
\State \textbf{Return} The global unlearned model $\theta_\lnot$;
\end{algorithmic}
\end{algorithm}

\subsection{Memory Generator}

The goal of the memory generator is to produce knowledge-free and easy-to-learn new memories for overwriting. 
In this paper, we focus on supervised learning, where the knowledge usually lies in labels.
Therefore, we manipulate labels to generate new memories.

\subsubsection{Knowledge-Free}

The primary characteristic of new memories is knowledge-free.
Removing the influence of the target data means making the model behave as if it has never seen the target data before.
In other words, the model is supposed to have no knowledge about the target data.
Following the idea of active forgetting, we generate knowledge-free labels, and overwrite the previously learned knowledge by training the model on these manipulated data (original features with new labels).

Assuming there is a binary classification task, and the label of a target data point can be $[0, 1]^\top$, which means it belongs to the second class.
There are two types of straightforward knowledge-free labels: i) the uniform label, e.g., $[0.5, 0.5]^\top$, and ii) the random label, e.g., $[r\sim\mathcal{U}(0, 1), 1 - r]^\top$ where $r$ is a random value sampled from a uniform distribution $\mathcal{U}(0, 1)$.
We compared with both types of the above labels in our ablation study (Section~\ref{sec:overlap}).

\subsubsection{Easy-to-Learn}

The above two introduced knowledge-free labels do not take prior knowledge of models into consideration, which makes them hard to learn by the model.
Our empirical study (Section~\ref{sec:overlap}) shows that both the uniform and random labels have residual old memories and unstable performance.
These hard-to-learn labels cause two potential problems during unlearning: (i) the incapacity of completely removing the memory of target data, and (ii) increasing the computational overhead for the unlearning process.

To make it easier to learn, we distill prior knowledge into labels using teacher-student learning paradigm~\citep{hinton2015distilling}.
We first randomly initialize a set of untrained target models as teacher models, as they have no learned knowledge about training data.
Then we feed teacher models with features of target data, producing the teacher predictions.
Finally, the teacher label is obtained by averaging all teacher predictions.
Formally, the teacher label $\hat{y}$ is computed as
\begin{equation}\label{equ:tea}
    \hat{y} = \frac{1}{Q}\sum_{i=1}^Q \theta_i(x),
\end{equation}
where $x$ is data features and $Q$ is the number of teachers.
Note that the teacher models are training-free and can be released after generating the fake labels.
Thus, the memory generator has both space and time advantages over the space-for-time strategy in existing retrain unlearning methods.

However, the prior knowledge can be so strong that it results in bias on particular data.
The bias is added purely by the model or algorithm, which is referred to as algorithmic bias~\citep{mehrabi2021survey}.
Our empirical study (Figure~\ref{fig:overlap}) shows that the untrained model naturally has prediction bias on particular classes, which makes the testing accuracy of these classes relatively high.
This potentially leaks knowledge about the data, which is conflicted with the primary characteristic, i.e., knowledge-free. 
Consequently, we import a debias vector $\nu$ to underweight the target class~\citep{espin2021explaining}.
Each element of $\nu$ represents the propensity score $\nu_i$ of the corresponding class.
We set the propensity score of target class $\nu_{\text{target}} = \sigma \in [0, 1]$ and the other classes $\nu_{\text{non-target}} = 1$.
Formally, the debias teacher label is computed as
\begin{equation}\label{equ:debias}
    \tilde{y} = \text{debias}(\hat{y}) = \frac{\nu \hat{y}}{|\nu \hat{y}|_1}, \enspace \nu = \boldsymbol{1} + (\sigma - 1)y,
\end{equation}
where $\boldsymbol{1}$ is an all-one vector, $y$ is the original label, and $\sigma$ is debias weight.
$\nu\hat{y}$ is scaled by its L1-norm to satisfy the summation constraint, i.e., $\sum y = 1$.
The debias weight can be dynamically determined by the quotient of the target prediction weight and average prediction weight.
In this way, we enjoy a balanced trade-off between the characteristics of knowledge-free and easy-to-learn.

\subsection{Knowledge Preserver}\label{sec:know}

While unlearning target data, we also need to preserve the remaining knowledge learned from non-target data.
Conventional deep learning methods suffer from the phenomenon of catastrophic forgetting when incrementally learning a sequence of tasks.
To imitate active forgetting in the human brain, our proposed \model{} is based on the idea of incremental learning.
The target client $k$ sequentially trains the model on two tasks, i.e., $\mathcal{D}_k$ and $\mathcal{M}$.
To avoid confusion, we directly name the task by its dataset.
Tasks $\mathcal{D}_k$ and $\mathcal{M}$ stand for the learning and unlearning process, respectively.
Catastrophic forgetting happens when training on the latter task, i.e., $\mathcal{M}$.
The new memories not only overwrite the old memories of target data, but also make the model forget the memories of non-target data.
Thus, we introduce refined EWC training to alleviate this phenomenon.

As introduced in Section~\ref{sec:over}, we do not change the learning process, which means that task $\mathcal{D}_k$ adopts conventional training.
The client $k$ optimizes model $\theta$ by minimizing the local loss $L_{\mathcal{D}_k}$, where we omit $\theta$ for simplicity.
At the end of training, $\theta$ is supposed to converge to a solution space:
\begin{equation}
    \theta^*_{\mathcal{D}_k} = \arg\min L_{\mathcal{D}_k},
\end{equation}
where $\forall \theta \in \theta^*_{\mathcal{D}_k}$ is an acceptable solution for task $\mathcal{D}_k$.

The next step is the unlearning process, where we train on task $\mathcal{M}$.
We explain the difference between conventional training and EWC training as follows.

\begin{figure}[t]
    \centering
    \includegraphics[width=0.7\linewidth]{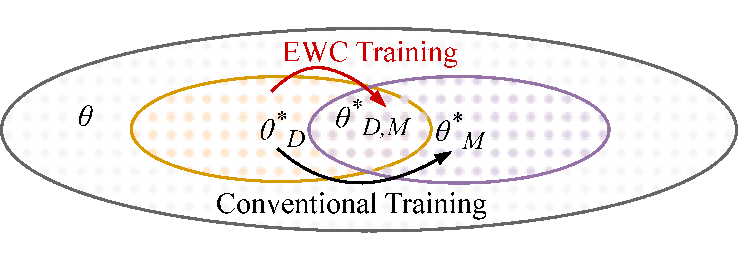}
    \caption{The difference between conventional training and EWC training. 
    The outermost circle denotes the space of model $\theta$.
    Assuming that the model is trained on tasks $\mathcal{D}_k$ and $\mathcal{M}$ sequentially.
    The circles of $\theta^*_{\mathcal{D}_k}$ and $\theta^*_\mathcal{M}$ denote possible solution spaces for task $\mathcal{D}_k$ and $\mathcal{M}$ respectively.
    Conventional training picks up the best possible parameters for task $\mathcal{M}$, i.e., $\theta^*_\mathcal{M}$, which leads to catastrophic forgetting of task $\mathcal{D}_k$.
    EWC training elastically regularizes the parameters on both tasks, and picks up the parameters in the overlapping solution space, i.e., $\theta^*_{\mathcal{D}_k \mathcal{M}}$.}
    \label{fig:ewc}
\end{figure}

\nosection{Conventional Training} Model $\theta^*_{\mathcal{D}_k}$ continues training by minimizing the loss of task $\mathcal{M}$, i.e., $L_\mathcal{M}$.
As shown in Figure~\ref{fig:ewc}, conventional training suffers from catastrophic forgetting, which is detrimental to model performance.

\nosection{EWC Training} A straightforward way to preserve knowledge is to train the model on both tasks simultaneously.
As shown in Figure~\ref{fig:ewc}, this method is based on the assumption that there is an \textit{overlapping solution space} of both tasks.
We empirically validate this assumption in Section~\ref{sec:overlap}.
Ignoring the scaling term $1/|\mathcal{D}', \mathcal{M}|$, the overall loss on both tasks is defined as
\begin{equation}
    L_{\mathcal{D}', \mathcal{M}} = L_\mathcal{M} + L_{\mathcal{D}'},
\end{equation}
where $\mathcal{D}' = \mathcal{D}_k \backslash \mathcal{R}$ denotes the task of non-target data, since we are required to unlearn the target data $\mathcal{R}$.
However, computing the loss of $\mathcal{D}'$ requires non-target data, which is against our intention of avoiding extra computational overhead.
EWC provides us an approximation of $L_{\mathcal{D}_k}$ when having $\theta^*_{\mathcal{D}_k}$, but not having access to $\mathcal{D}_k$.
The overall loss $L_{\mathcal{D}_k, \mathcal{M}}$ is approximated by computing the second order Taylor expansion of $L_{\mathcal{D}_k}$ at $\theta^*_{\mathcal{D}_k}$ as follows:
\begin{align}\label{equ:ewc}
    L_{\mathcal{D}_k, \mathcal{M}} & = L_\mathcal{M} + L_{\mathcal{D}_k}\\ \nonumber
    & \approx L_\mathcal{M} + \frac{\lambda}{2}(\theta - \theta^*_{\mathcal{D}_k})^\top H_{\mathcal{D}_k}(\theta - \theta^*_{\mathcal{D}_k}) + \epsilon,
\end{align}
where $\lambda$ is a hyper-parameter introduced to have a trade-off between learning $\mathcal{M}$ and not forgetting $\mathcal{D}_k$, $H_{\mathcal{D}_k} = \partial^2L(\theta^*_{\mathcal{D}_k})/\partial^2\theta^*_{\mathcal{D}_k}$ denotes Hessian matrix at $\theta^*_{\mathcal{D}_k}$, and $\epsilon$ accounts for all constants.
Note that we omit the first-order term of the Taylor expansion because the gradient $\partial L(\theta^*_{\mathcal{D}_k})/\partial \theta^*_{\mathcal{D}_k}$ is approximately zero, as $\theta^*_{\mathcal{D}_k} = \arg\min L_{\mathcal{D}_k}$.
Please refer to~\citep{kolouri2020sliced} for more details of derivation.
As shown in Eq~(\ref{equ:ewc}), Hessian matrix can be interpreted as regularizer weight matrix of $(\theta - \theta^*_{\mathcal{D}_k})$, where each element represents synaptic importance of a parameter.
The more important the parameter is, the more regularizer weight it should be added to, making the parameter change less when learning $\mathcal{M}$.

\nosection{Acceleration} 
Previous work computes the Hessian by Fisher information matrix~\citep{martens2020new}.
To enhance efficiency, we use sketch matrix~\citep{li2021lifelong} to approximate it.
Specifically, the Hessian $H_{\mathcal{D}_k} \in \mathbb{R}^{m\times m}$ has a natural decomposition~\citep{aljundi2018memory} as:
\begin{equation}
    H_{\mathcal{D}_k} = \frac{1}{n_k}J^{\top}J,
\end{equation}
where each row of $J \in \mathbb{R}^{n_k \times m}$ is a Jacobian matrix of a sample from $\mathcal{D}_k$.
Thus, we have
\begin{equation}
    \frac{\lambda}{2}(\theta - \theta_{\mathcal{D}_k}^*)^{\top}H_{\mathcal{D}_k}(\theta - \theta_{\mathcal{D}_k}^*) = \frac{\lambda}{2n_k}\|J\cdot (\theta - \theta_{\mathcal{D}_k}^*)\|_2^2.
\end{equation}
Consequently, we can rewrite Eq~(\ref{equ:ewc}) as:
\begin{equation}
    L_{\mathcal{D}_k, \mathcal{M}} \approx L_\mathcal{M} + \frac{\lambda}{2n_k}\|J\cdot(\theta - \theta^*_{\mathcal{D}_k})\|^2_2 + \epsilon.
\end{equation}
However, $J$ would also cost considerable computational overhead.
Thus, we approximate it with a sketch matrix as follows~\citep{charikar2002finding,li2021lifelong}:
\begin{equation}
    \hat{J} = S J \in \mathbb{R}^{s\times m},
\end{equation}
where $S \in \mathbb{R}^{s\times n_k}$ can be interpreted as sampled from a carefully chosen distribution, and $s \ll \min\{n_k, m\}$. 
\begin{theorem}[Error Upper Bound~\citep{li2021lifelong}]\label{the:bound}
    For a matrix $J \in \mathbb{R}^{n_k \times m}$ with stable rank $r$, a transformation matrix $S\in \mathbb{R}^{s\times n_k}$ with $s = O(r^2/\varepsilon^2)$ has the property that for all $\theta \in \mathbb{R}^m$, 
    \begin{equation}
        |\|SJ\theta\|_2^2 - \|J\theta\|_2^2| \le \varepsilon \|J\|_2^2\cdot\|\theta\|_2^2,
    \end{equation}
    with probability at least 0.99, where $r = \|J\|_F^2 / \|J\|_2^2$.
\end{theorem}
Theorem~\ref{the:bound} reveals that this approximation is bounded by the scale of $\|J\|_F^2$, which can be interpreted as the scale of the gradient. The derivation is formally presented in Algorithm~\ref{alg:ske}.

\begin{algorithm}[tb]
\caption{Sketch Matrix Construction}
\label{alg:ske}
\textbf{Require}: Jocobian matrix $J$, samples in $\mathcal{D}_k$ and sketch size $s \in \mathbb{N}^+$.\\
\textbf{Ensure}: 2-wise and 4-wise independent hash functions $h: [n_k] \to [s]$ and $\sigma: [n_k] \to \{1, -1\}$ respectively.\\
\textbf{Procedure}:
\begin{algorithmic}[1] 
\For{$i = 1, \dots, s$}
    \State Group data $G_i := \{x\in \mathcal{D}_k: h(x)=i\}$;
    \State Set $(\hat{J})_i \gets \sum_{x\in G_i} \sigma(x)(J)_x$;
\EndFor
\State \textbf{return} $\hat{J} \in \mathbb{R}^{s\times m}$;
\end{algorithmic}
\end{algorithm}

Compared with other low-rank approximate approaches, e.g., PCA~\citep{bro2014principal}, this sketch-based approach provides a small approximate error with only a minimal amount of additional computation.

\nosection{EWC Unlearning}
Based on the above approximation of $L_{\mathcal{D}_k}$, we compute the overall loss on tasks $\mathcal{D}'$ and $\mathcal{M}$, i.e., the unlearning loss, as follows:
\begin{align}\label{equ:new}
    & L_{\mathcal{D}', \mathcal{M}} = L_\mathcal{M} + L_{\mathcal{D}'} = L_\mathcal{M} + L_{\mathcal{D}_k} - L_\mathcal{R}\\ \nonumber
    & \approx L_\mathcal{M} - L_\mathcal{R} + \frac{\lambda}{2}\|\hat{J}\cdot(\theta - \theta^*_{\mathcal{D}_k})\|^2_2 + \epsilon,
\end{align}
where $\mathcal{R}$ and $\mathcal{M}$ denote the original target data (old memories) and the manipulated target data (new memories) respectively.
Since both $n_k$ and $\lambda$ are scales, we can omit $n_k$ without loss of generality.
Consequently, the model $\theta^*_{\mathcal{D}_k}$ continues training on our proposed new loss to alleviate the phenomenon of catastrophic forgetting.
In this way, the knowledge preserver can unlearn the target data without breaking the original training procedure.
\lyy{As shown in Theorem~\ref{the:bound}, the update process of EWC unlearning is bounded by the scale of the gradient, meaning it is computationally equivalent to the learning process in terms of speed. Additionally, our proposed framework only requires continuous fine-tuning on the target data (typically a small portion of the entire dataset), making it theoretically much more efficient than retraining from scratch in the federated setting.}

\section{Experiments}

We evaluate the effectiveness of our proposed method, \model{}, across three widely used datasets based on the following three principles, i.e., unlearning completeness, unlearning efficiency, and model utility. 
Note that \model{} inherently adheres to the principle of data privacy by virtue of its design, which solely involves local data manipulation at the client side.
In addition, we conduct an overlapping validation experiment to provide fundamental support for EWC training.
To further investigate our proposed method, we also conduct an ablation study of two modules in our proposed method and analyze parameter sensitivity.

\subsection{Experimental Settings}

\subsubsection{Datasets}
We conduct experiments on three benchmark datasets, i.e., MNIST~\citep{lecun1998gradient}, CIFAR10~\citep{krizhevsky2009learning}, and CelebA~\citep{liu2015faceattributes}. We provide the statistics of these datasets in Table~\ref{tab:dataset}.
For MNIST and CIFAR10, following their original papers, we leave 10,000 samples for testing and use the others for training.
For CelebA, we select two representative attributes, i.e., \textit{Male} and \textit{Mouth\_Slightly\_Open}. 
We label the samples based on whether they possess these attributes or not, which results in four distinct classes, i.e., having both attributes, having none of the attributes, only having \textit{Male} attribute, and only having \textit{Mouth\_Slightly\_Open} attribute.
We use 80\%, 10\%, and 10\% of the original dataset for training, validation, and testing, respectively.

We use a network that consists of 2 convolutional layers followed by 1 fully-connected layer for MNIST, ResNet-10~\citep{he2016deep} for CIFAR10 and CelebA.
For simplicity, we did not sufficiently fine-tune the models to get their optimal performance, which is not the focus of this paper.
Both networks can achieve over 80\% accuracy on testing set during learning.

\begin{table}
    \centering
    \caption{Summary of datasets.}
    \begin{tabular}{cccc}
        \toprule
        Dataset & Dimensionality & Size & \# Classes \\
        \midrule
        MNIST   & 28 $\times$ 28            & 70,000 & 10\\
        CIFAR10 & 32 $\times$ 32 $\times$ 3 & 60,000 & 10\\
        CelebA  & 178 $\times$ 218 $\times$ 3 & 202,599 & 4\\
        \bottomrule
    \end{tabular}
    \label{tab:dataset}
\end{table}

\subsubsection{Training Details} We initialize all model parameters, including teacher models, with Xavier Initialization~\citep{glorot2010understanding}.
As initialization involves randomness, we run all models for 10 trials and report the average results.
We adopt the cross entropy loss function and stochastic gradient descent to train the models.
The learning rate is set as 0.001 for MNIST, 0.01 for CIFAR10, and 0.01 for CelebA.
In this paper, we conduct class-wise unlearning, and unlearn one class of a specific client at one time for all classes.
All models are implemented with PyTorch and ran on an Ubuntu 20.04 LTS system server with 256GB RAM, and NVIDIA GeForce RTX 3090 GPU.
We run all experiments for 10 trials and report the average results.

\subsubsection{Federated Settings} For federated learning, we adopt the widely acknowledged FedAvg~\citep{mcmahan2017communication}.
Specifically, we set the number of clients as 4, and the local epoch as 1.
Ensuring convergence of the global model, the maximum round of federated iteration is set as 5, 20, and 10 for MNIST, CIFAR10, and CelebA, respectively.

\subsubsection{Hyper-parameters} In our proposed \model{}, there are three hyper-parameters, i.e., EWC training epoch, trade-off coefficient $\lambda$, and the number of teacher models $Q$.
For the EWC training epoch, we investigate it in \{1, 2, 3, 4, 5\}.
For $\lambda$, we investigate it in \{0.1, 0.5, 1, 10, 50\}.
Based on empirical results, we set the EWC training epoch as 1 and $\lambda$ as 10 for the following experiments in this paper.
We report empirical results in Section~\ref{sec:para}.
For $Q$, we investigate it in \{5, 10, 15, 20\} and finally set $Q=10$, since we observe that a larger number over 10 cannot significantly improve the overall performance of teacher models.

\begin{figure}[t]
    \centering
    \subfigure[MNIST: Ori]{
        \includegraphics[width=0.14\textwidth]{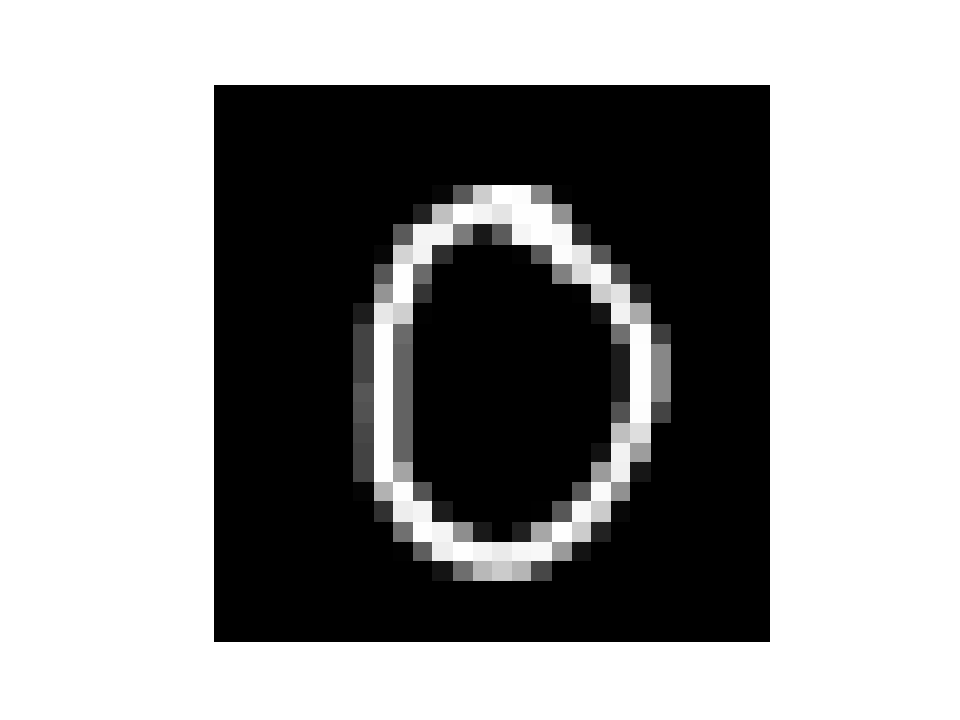}
    }
    \subfigure[CIFAR10: Ori]{
        \includegraphics[width=0.14\textwidth]{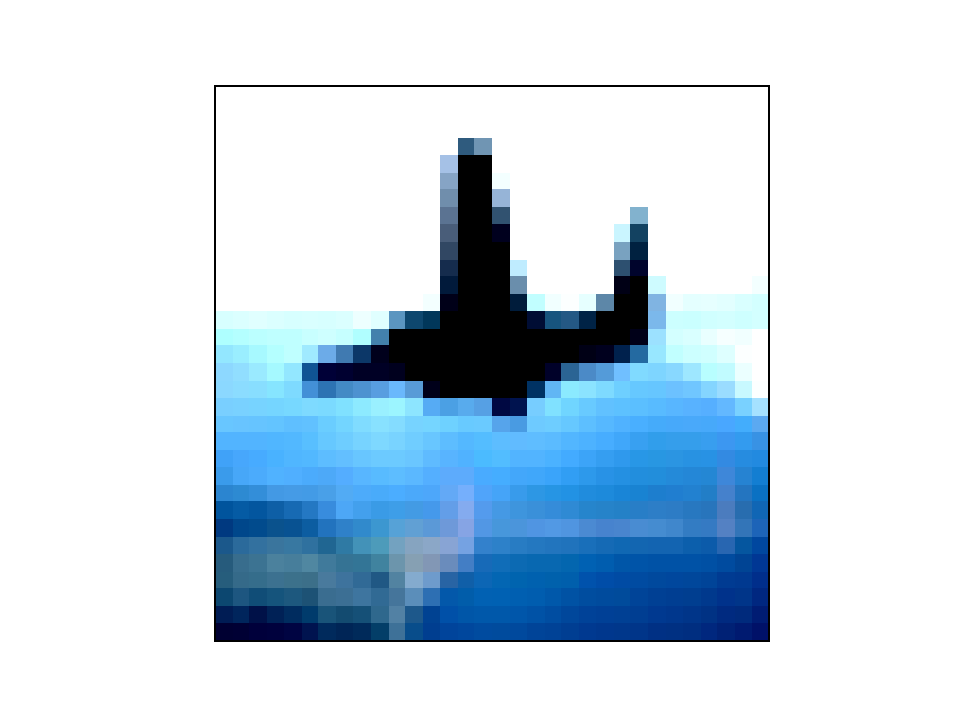}
    }
    \subfigure[CelebA: Ori]{
        \includegraphics[width=0.14\textwidth]{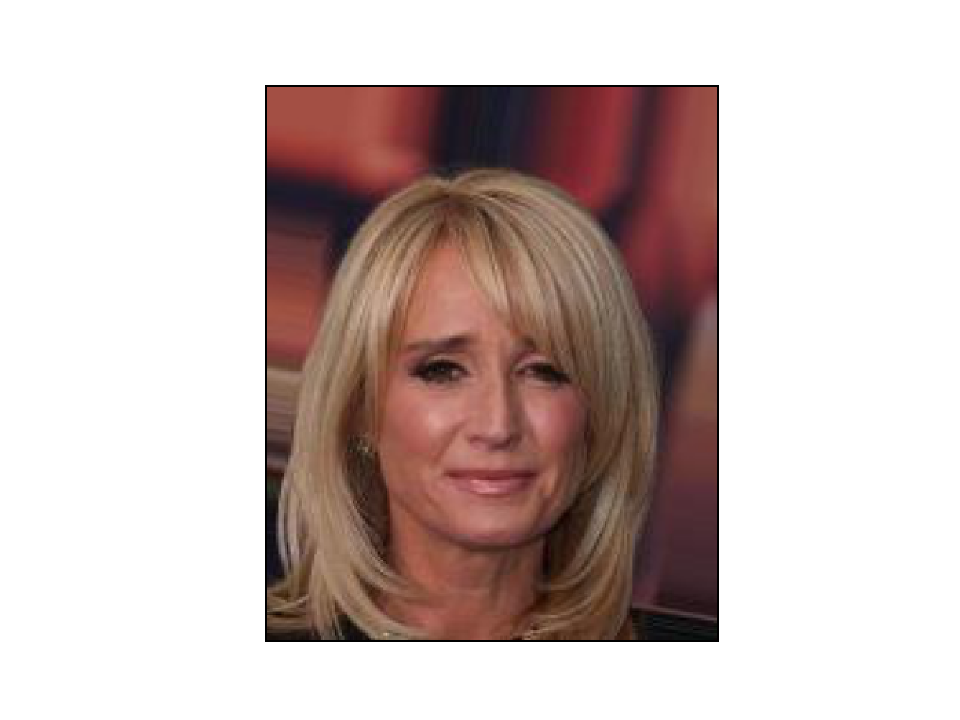}
    }
    \subfigure[MNIST: BKD]{
        \includegraphics[width=0.14\textwidth]{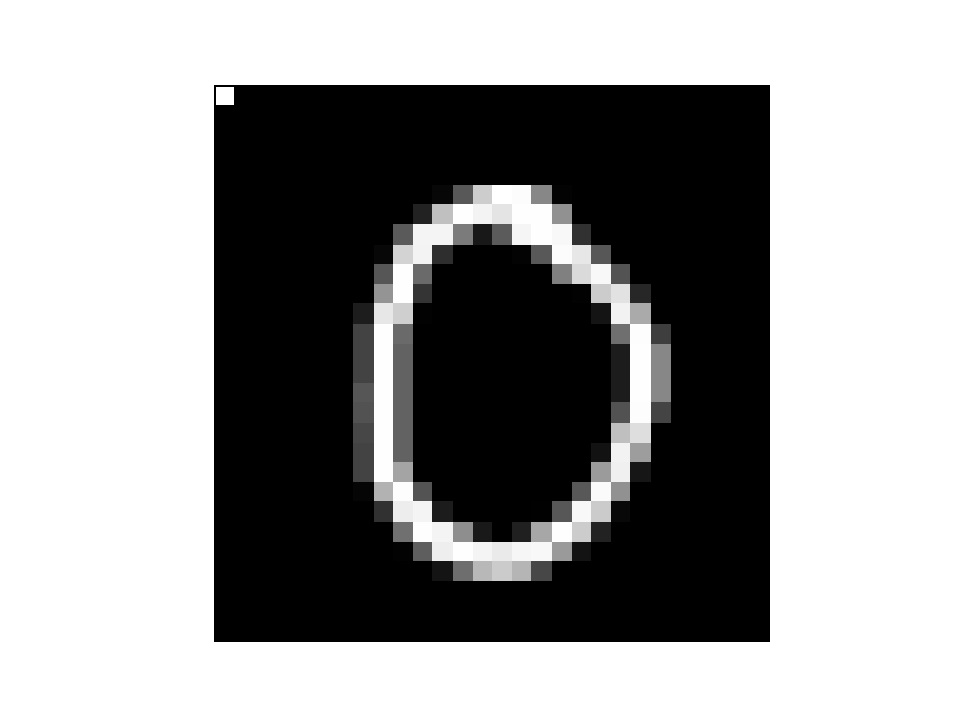}
    }
    \subfigure[CIFAR10: BKD]{
        \includegraphics[width=0.14\textwidth]{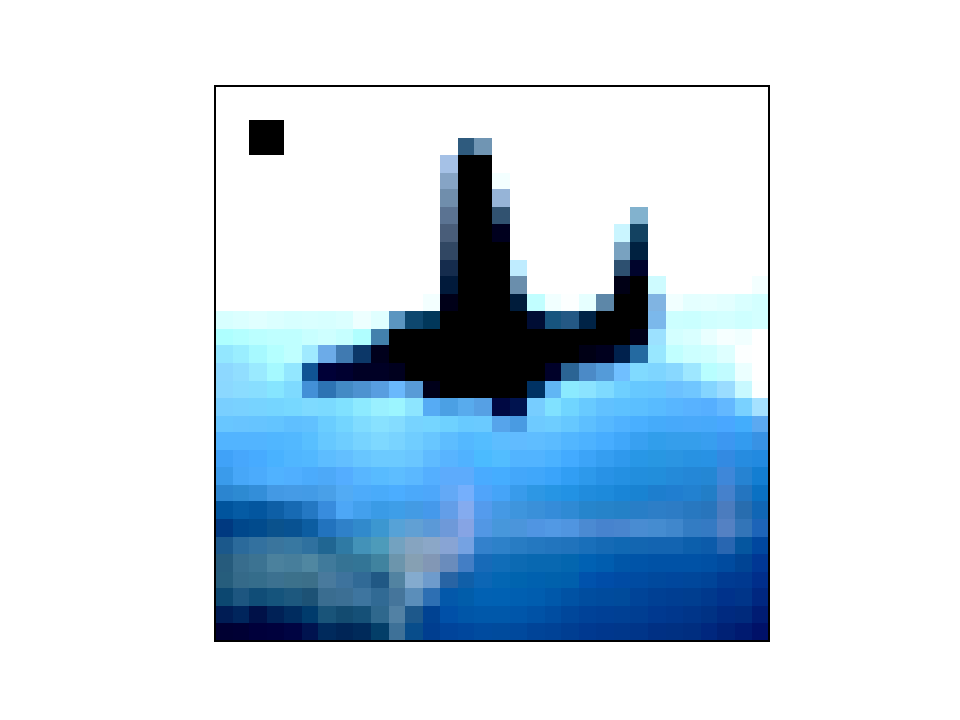}
    }
    \subfigure[CelebA: BKD]{
        \includegraphics[width=0.14\textwidth]{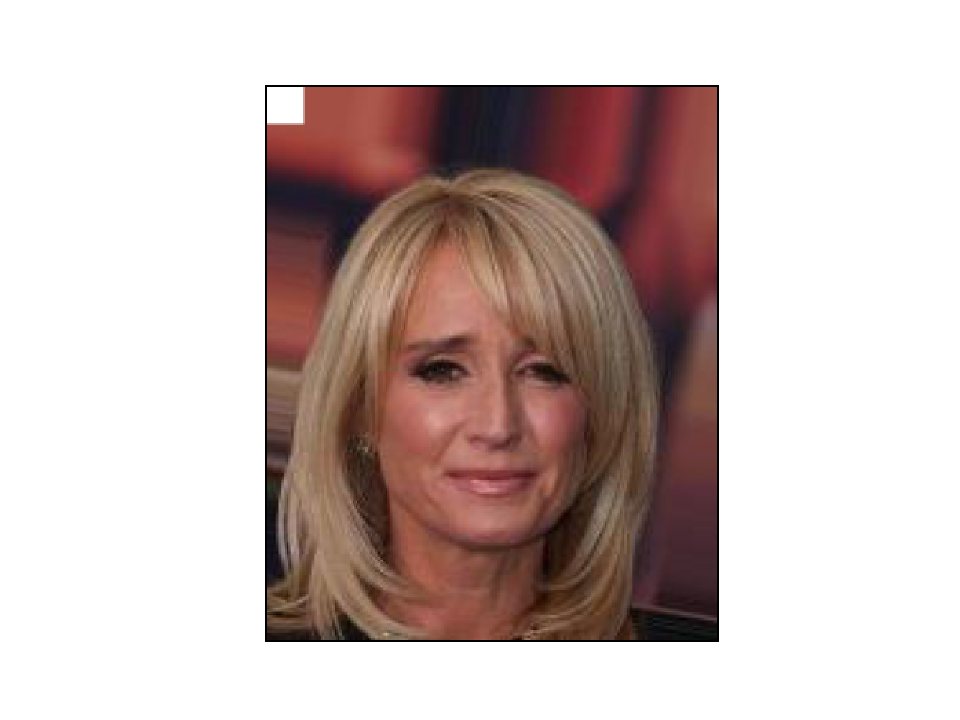}
    }
    \caption{Illustrations of original images (Ori, i.e., the first row) in the datasets, and the images that are implanted with backdoor triggers (BKD, i.e., the second row).}
    \label{fig:backdoor}
\end{figure}

\begin{table*}[t]
\centering
\caption{Accuracy (Acc) of unlearning different classes.
The BD Acc (the lower the better) and the Test Acc are used to evaluate unlearning completeness and model utility, respectively.
The accuracy of testing data (Test Acc) before unlearning is (i) MNIST: 98.30 $\pm$ 0.21, (ii) CIFAR10: 82.04 $\pm$ 0.36, and (iii) CelebA: 90.31 $\pm$ 0.22.
We \textbf{bold} the top results of each class, other than Retrain.}
\resizebox{\textwidth}{!}{
\begin{tabular}{c|c|cccccc|cccccc}  
\toprule
\multirow{2}{*}{Class ID} & \multicolumn{1}{c|}{BD Acc (\%)} & \multicolumn{6}{c}{BD Acc (\%) after unlearning $\downarrow$} & \multicolumn{6}{c}{Test Acc (\%) after unlearning $\uparrow$}\\
    & before unlearning & Retrain & FRR & SGA & CDP & \model{}-C & \model{} & Retrain & FRR & SGA & CDP & \model{}-C & \model{}\\
\midrule
\multicolumn{14}{c}{MNIST}\\
\midrule
0 & 81.14$\pm$2.86 & 0.12$\pm$0.01 & 0.73$\pm$0.13 & 2.71$\pm$0.94 & 65.43$\pm$1.12 & 0.60$\pm$0.35 & \textbf{0.58$\pm$0.18} & 96.83$\pm$0.13 & 91.14$\pm$0.12 & 71.32$\pm$2.21 & 84.39$\pm$1.13 & 40.03$\pm$0.51 & \textbf{92.22$\pm$0.26}\\
1 & 81.57$\pm$3.32 & 0.11$\pm$0.14 & 0.55$\pm$0.08 & \textbf{0.00$\pm$0.00} & 70.04$\pm$1.54 & 0.68$\pm$0.52 & 0.82$\pm$0.43 & 96.84$\pm$0.07 & 91.06$\pm$0.12 & 69.27$\pm$0.98 & 80.17$\pm$1.45 & 39.99$\pm$0.41 & \textbf{92.29$\pm$0.61}\\
2 & 81.58$\pm$3.29 & 0.03$\pm$0.00 & 0.94$\pm$0.15 & \textbf{0.00$\pm$0.00} & 71.07$\pm$0.99 & 0.40$\pm$0.35 & 0.65$\pm$0.41 & 96.39$\pm$0.18 & 90.75$\pm$0.20 & 65.45$\pm$2.07 & 85.11$\pm$1.28 & 39.75$\pm$0.32 & \textbf{92.06$\pm$0.39}\\
3 & 80.73$\pm$3.19 & 0.08$\pm$0.01 & 0.62$\pm$0.04 & \textbf{0.00$\pm$0.00} & 67.22$\pm$1.32 & 0.71$\pm$0.58 & 0.79$\pm$0.45 & 96.68$\pm$0.07 & 90.42$\pm$0.28 & 70.33$\pm$0.87 & 84.09$\pm$1.10 & 39.94$\pm$0.33 & \textbf{92.19$\pm$0.38}\\
4 & 81.16$\pm$2.87 & 0.01$\pm$0.00 & 0.60$\pm$0.02 & 1.79$\pm$0.88 & 69.37$\pm$1.35 & 0.93$\pm$0.18 & \textbf{0.39$\pm$0.05} & 96.42$\pm$0.10 & 90.52$\pm$0.15 & 68.88$\pm$1.17 & 84.11$\pm$0.94 & 39.92$\pm$0.62 & \textbf{92.12$\pm$0.64}\\
5 & 80.82$\pm$3.13 & 0.37$\pm$0.15 & 0.97$\pm$0.20 & 2.01$\pm$1.15 & 73.22$\pm$1.52 & 0.62$\pm$0.41 & \textbf{0.57$\pm$0.32} & 96.78$\pm$0.13 & 90.42$\pm$0.22 & 70.33$\pm$0.91 & 80.77$\pm$1.25 & 40.10$\pm$0.24 & \textbf{92.12$\pm$0.62}\\
6 & 80.25$\pm$4.06 & 0.04$\pm$0.00 & 1.04$\pm$0.17 & \textbf{0.00$\pm$0.00} & 66.37$\pm$1.32 & 0.55$\pm$0.07 & 0.73$\pm$0.20 & 96.74$\pm$0.05 & 91.03$\pm$0.21 & 71.55$\pm$0.95 & 87.15$\pm$1.08 & 40.17$\pm$0.47 & \textbf{91.85$\pm$0.60}\\
7 & 80.41$\pm$3.50 & 0.27$\pm$0.15 & 1.71$\pm$0.54 & \textbf{0.00$\pm$0.00} & 70.23$\pm$1.23 & 1.29$\pm$0.45 & 1.27$\pm$0.62 & 96.54$\pm$0.15 & 91.01$\pm$0.14 & 68.35$\pm$0.77 & 83.89$\pm$0.95 & 39.94$\pm$0.45 & \textbf{92.04$\pm$0.24}\\
8 & 81.86$\pm$3.57 & 0.13$\pm$0.04 & \textbf{0.77$\pm$0.21} & 1.99$\pm$0.71 & 69.42$\pm$1.04 & 0.43$\pm$0.10 & 0.82$\pm$0.18 & 96.68$\pm$0.24 & 90.21$\pm$0.22 & 69.57$\pm$0.94 & 82.25$\pm$0.75 & 40.11$\pm$0.71 & \textbf{91.90$\pm$0.42}\\
9 & 79.60$\pm$5.30 & 0.22$\pm$0.06 & 1.41$\pm$0.47 & 2.35$\pm$0.46 & 69.49$\pm$0.74 & 1.33$\pm$0.22 & \textbf{1.05$\pm$0.50} & 96.38$\pm$0.21 & 89.76$\pm$0.34 & 68.43$\pm$1.68 & 84.35$\pm$0.85 & 40.07$\pm$0.37 & \textbf{92.38$\pm$0.65}\\
\midrule
\multicolumn{14}{c}{CIFAR10}\\
\midrule
0 & 71.00$\pm$1.73 & 0.01$\pm$0.00 & 1.64$\pm$1.05 & \textbf{0.00$\pm$0.00} & 56.35$\pm$1.33 & 1.45$\pm$1.87 & 1.45$\pm$1.02 & 81.03$\pm$0.63 & \textbf{80.04$\pm$0.61} & 61.82$\pm$1.19 & 72.46$\pm$1.06 & 29.91$\pm$6.74 & 79.55$\pm$1.92\\
1 & 70.59$\pm$4.62 & 1.63$\pm$1.28 & 2.38$\pm$1.91 & \textbf{0.00$\pm$0.00} & 55.02$\pm$1.91 & 2.17$\pm$1.44 & 2.81$\pm$1.80 & 80.84$\pm$0.27 & 79.04$\pm$0.31 & 63.73$\pm$1.22 & 72.65$\pm$0.81 & 28.27$\pm$4.11 & \textbf{79.76$\pm$0.96}\\
2 & 74.17$\pm$4.02 & 0.86$\pm$0.08 & 1.88$\pm$0.91 & \textbf{0.00$\pm$0.00} & 55.42$\pm$1.70 & 1.50$\pm$0.48 & 1.20$\pm$0.91 & 81.78$\pm$0.57 & \textbf{81.50$\pm$0.56} & 68.07$\pm$0.92 & 73.61$\pm$0.51 & 27.87$\pm$3.60 & 80.82$\pm$0.75\\
3 & 69.85$\pm$1.29 & 1.59$\pm$1.25 & 2.15$\pm$1.71 & \textbf{0.00$\pm$0.00} & 67.06$\pm$2.01 & 1.79$\pm$1.10 & 1.76$\pm$0.83 & 81.61$\pm$1.06 & 80.60$\pm$1.09 & 68.85$\pm$1.01 & 73.48$\pm$1.02 & 28.18$\pm$8.25 & \textbf{80.63$\pm$1.76}\\
4 & 70.55$\pm$4.44 & 0.06$\pm$0.00 & 1.34$\pm$1.04 & 1.79$\pm$1.64 & 61.84$\pm$0.88 & 1.55$\pm$0.95 & \textbf{1.30$\pm$1.07} & 81.25$\pm$0.40 & 80.15$\pm$0.35 & 68.53$\pm$0.88 & 73.39$\pm$0.94 & 27.07$\pm$4.61 & \textbf{80.27$\pm$0.81}\\
5 & 74.66$\pm$1.07 & 0.03$\pm$0.01 & 1.34$\pm$1.03 & \textbf{1.14$\pm$1.11} & 64.25$\pm$1.44 & 1.96$\pm$1.24 & 1.70$\pm$1.12 & 80.57$\pm$0.25 & 79.59$\pm$0.25 & 68.50$\pm$1.63 & 73.63$\pm$1.02 & 27.85$\pm$5.35 & \textbf{80.50$\pm$1.74}\\
6 & 70.38$\pm$2.42 & 0.09$\pm$0.01 & 1.30$\pm$0.94 & 2.21$\pm$0.77 & 55.02$\pm$0.88 & 1.76$\pm$1.19 & \textbf{1.28$\pm$1.06} & 81.17$\pm$0.81 & 80.18$\pm$0.84 & 63.09$\pm$2.15 & 72.80$\pm$0.77 & 30.19$\pm$4.42 & \textbf{81.05$\pm$1.02}\\
7 & 73.66$\pm$6.68 & 0.05$\pm$0.00 & 1.34$\pm$1.05 & \textbf{0.00$\pm$0.00} & 53.41$\pm$1.11 & 1.56$\pm$1.12 & 1.53$\pm$1.22 & 80.90$\pm$0.42 & 80.44$\pm$0.41 & 63.99$\pm$1.54 & 72.52$\pm$0.37 & 25.45$\pm$2.42 & \textbf{80.78$\pm$0.50}\\
8 & 79.05$\pm$4.81 & 0.02$\pm$0.00 & 1.54$\pm$0.92 & 2.41$\pm$0.75 & 54.61$\pm$0.93 & 1.84$\pm$0.85 & \textbf{1.41$\pm$0.66} & 80.71$\pm$0.29 & 79.68$\pm$0.33 & 70.51$\pm$1.96 & 72.36$\pm$0.41 & 29.14$\pm$3.48 & \textbf{80.37$\pm$0.25}\\
9 & 81.70$\pm$6.54 & 1.08$\pm$0.07 & 1.81$\pm$1.07 & 2.81$\pm$1.40 & 60.64$\pm$1.02 & 1.75$\pm$1.46 & \textbf{1.55$\pm$1.15} & 80.54$\pm$0.57 & 79.45$\pm$0.64 & 69.95$\pm$0.98 & 72.57$\pm$1.10 & 29.14$\pm$2.67 & \textbf{80.28$\pm$1.12}\\
\midrule
\multicolumn{14}{c}{CelebA}\\
\midrule
0 & 74.16$\pm$0.37 & 0.68$\pm$0.30 & 2.18$\pm$1.30 & \textbf{0.00$\pm$0.00} & 64.56$\pm$0.27 & 1.76$\pm$1.31 & 1.91$\pm$1.32 & 88.71$\pm$0.36 & \textbf{88.69$\pm$0.39} & 32.07$\pm$3.85 & 87.18$\pm$0.22 & 31.29$\pm$4.91 & 88.44$\pm$0.47\\
1 & 74.58$\pm$1.65 & 0.39$\pm$0.35 & 1.43$\pm$0.86 & \textbf{0.00$\pm$0.00} & 46.06$\pm$0.77 & 1.29$\pm$1.14 & 1.34$\pm$1.15 & 89.29$\pm$0.24 & 89.09$\pm$0.26 & 32.76$\pm$1.83 & 87.91$\pm$1.10 & 32.35$\pm$2.73 & \textbf{89.10$\pm$1.18}\\
2 & 84.55$\pm$3.40 & 1.33$\pm$0.61 & 2.58$\pm$1.44 & \textbf{0.00$\pm$0.00} & 46.28$\pm$1.47 & 2.52$\pm$2.26 & 2.43$\pm$1.65 & 86.15$\pm$0.25 & \textbf{86.10$\pm$0.29} & 34.39$\pm$3.04 & 83.68$\pm$1.41 & 29.99$\pm$4.11 & 85.14$\pm$0.31\\
3 & 62.13$\pm$6.33 & 0.63$\pm$0.80 & 1.94$\pm$1.31 & \textbf{0.00$\pm$0.00} & 36.43$\pm$0.98 & 1.82$\pm$1.24 & 1.98$\pm$1.73 & 89.34$\pm$0.37 & 88.11$\pm$0.20 & 31.53$\pm$3.22 & 87.57$\pm$0.82 & 29.83$\pm$3.94 & \textbf{89.74$\pm$0.56}\\
\bottomrule
\end{tabular}
}
\label{tab:complete}
\end{table*}

\subsubsection{Compared Methods}
We compare our proposed \model{} with representative unlearning methods which are applicable for class-wise unlearning.
\begin{itemize}
    \item \textbf{Retrain}: Retraining from scratch is the ground-truth unlearning method with heavy computational overhead. 
    \item \textbf{FRR}~\citep{liu2022right}: Federated Rapid Retraining is a representative sample-wise federated unlearning method, which can be applied to class-wise unlearning.
    We implement FRR using the published code.
    \item \textbf{SGA}~\citep{wu2022federated}: Stochastic Gradient Ascent incorporates regularization to achieve unlearning in federated learning.
    \item \textbf{CDP}~\citep{wang2022federated}: Class-Discriminative Pruning is a class-wise federated unlearning method designed for CNN. In our experiments, CDP is applicable for ResNet-10 and convolutional layers in the model used for MNIST.
\end{itemize}

\subsection{Results and Discussions}

\subsubsection{Unlearning Completeness}
Following~\citep{sommer2022athena,hong2022membership}, we utilize backdoor attack~\citep{gu2019badnets} to evaluate the unlearning completeness.
Specifically, we implant backdoor triggers into the target data that we aim to unlearn, and flip their labels to a random class.
Through this process, the backdoor attack enforces the model to build a mapping between the trigger pattern and the flipped label, overwhelming the mapping between the original pattern and true label.
In Figure~\ref{fig:backdoor}, we provide several examples of images that have been implanted with backdoor triggers.
These backdoor triggers enable us to achieve an average attack success rate (i.e., BackDoor 
 Accuracy, BD Acc) of 80.91\%, 73.56\%, and 73.86\% for MNIST, CIFAR10, and CelebA, respectively.
As a consequence, we can evaluate unlearning completeness by comparing the performance of target data before and after unlearning.
We report BD Acc of the unlearned data in Table~\ref{tab:complete}.
From it, we observe that: 
\begin{itemize}
    \item Except CDP (61.20\% on average), all compared methods significantly decrease the BD Acc after unlearning (from 76.67\% to 1.07\% on average), indicating that the influence of target data is significantly reduced.
    \item FRR, SGA, and \model{} cannot achieve as low BD Acc (after unlearning) as Retrain, which suggests that all compared unlearning methods cannot achieve the same degree of unlearning completeness as Retrain. 
    Compared with the BD Acc of Retrain, FRR, SGA, and \model{} have an average increase of 2.48\%, 1.16\%, and 2.19\%, respectively.
    \item Although SGA demonstrates the second-best overall completeness performance (just below Retrain), it occasionally overkills, i.e., lower accuracy than Retrain, resulting in a 0.00\% BD Acc.
    This consequently leads to significantly reduced model utility, as we will illustrate in Section~\ref{sec:utility}. Overall, \model{} achieves satisfying performance in terms of unlearning completeness.
\end{itemize}

\begin{table}
    \centering
    \caption{Running time of unlearning one class in a client.}
    \resizebox{\linewidth}{!}{
    \begin{tabular}{c|rrrrr}
        \toprule
        Time (s) & Retrain  & FRR & SGA & CDP & FedAF \\
        \midrule
        MNIST   & 24.55     & 28.60 & 1.29 & 17.39 & 1.62\\
        CIFAR10 & 891.46     & 3,554.32 & 4.78 & 34.37 & 9.29\\
        CelebA  & 1,774.75     & 16,696.48 & 36.15 & 235.95 & 57.20\\
        \bottomrule
    \end{tabular}
    }
    \label{tab:time}
\end{table}

\begin{figure*}[t]
    \centering
    \subfigure[MNIST: Uniform]{
        \includegraphics[width=0.23\textwidth]{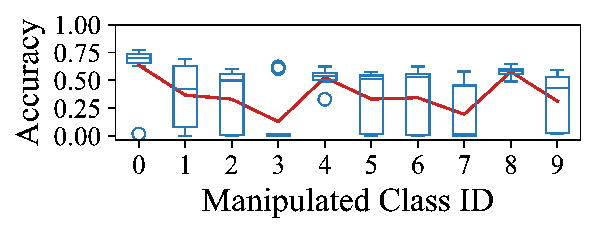}
    }
    \subfigure[MNIST: Random]{
        \includegraphics[width=0.23\textwidth]{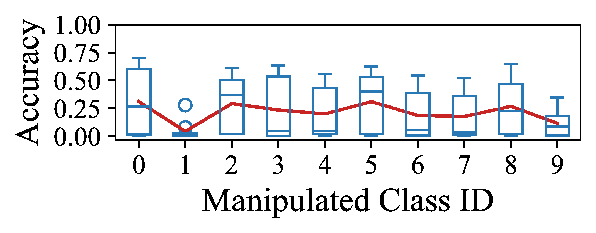}
    }
    \subfigure[MNIST: Teacher]{
        \includegraphics[width=0.23\textwidth]{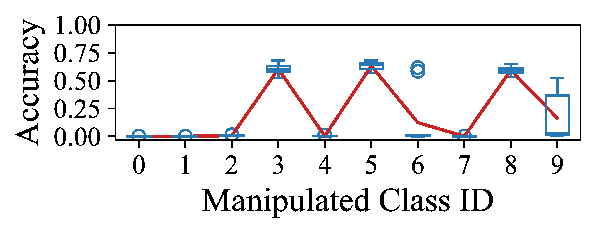}
    }
    \subfigure[MNIST: Debias Teacher]{
        \includegraphics[width=0.23\textwidth]{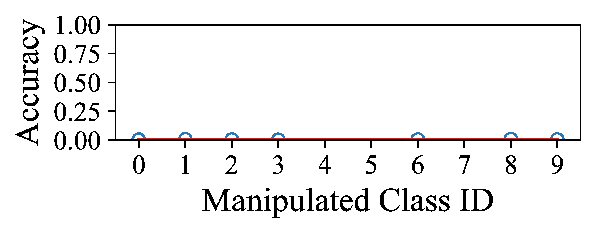}
    }
    \subfigure[CIFAR10: Uniform]{
        \includegraphics[width=0.23\textwidth]{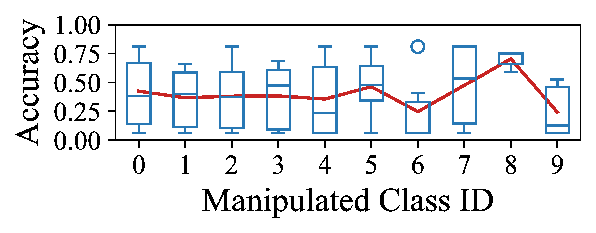}
    }
    \subfigure[CIFAR10: Random]{
        \includegraphics[width=0.23\textwidth]{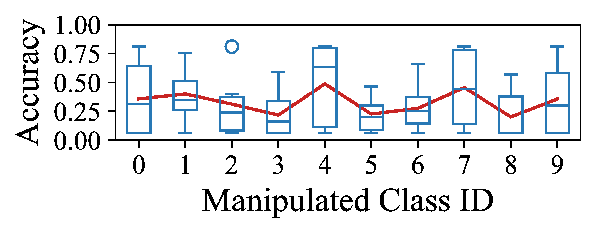}
    }
    \subfigure[CIFAR10: Teacher]{
        \includegraphics[width=0.23\textwidth]{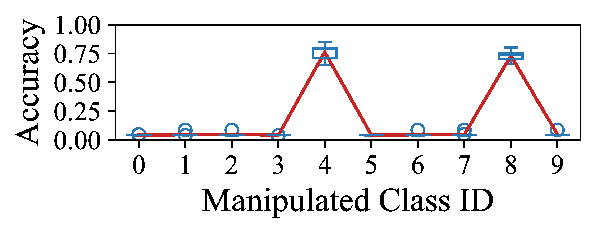}
    }
    \subfigure[CIFAR10: Debias Teacher]{
        \includegraphics[width=0.23\textwidth]{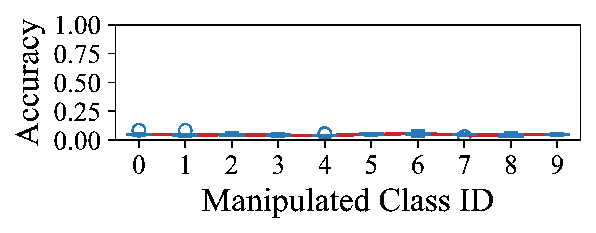}
    }
    \subfigure[CelebA: Uniform]{
        \includegraphics[width=0.23\textwidth]{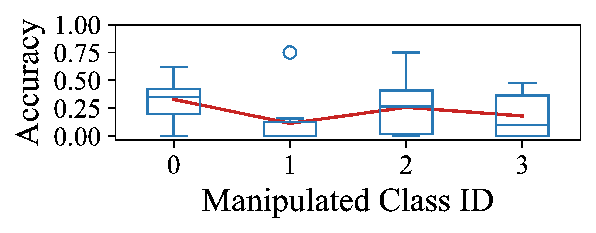}
    }
    \subfigure[CelebA: Random]{
        \includegraphics[width=0.23\textwidth]{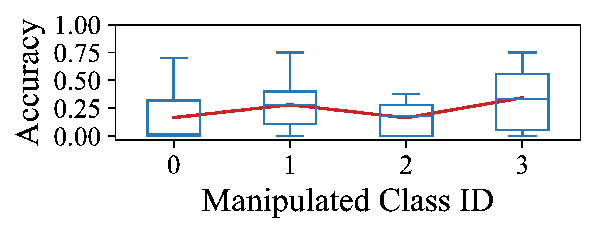}
    }
    \subfigure[CelebA: Teacher]{
        \includegraphics[width=0.23\textwidth]{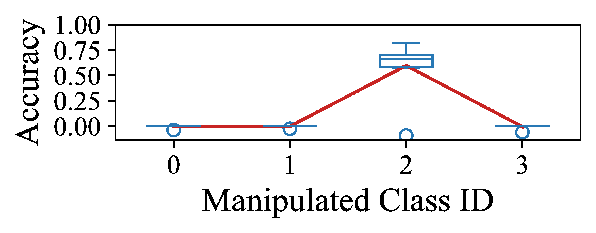}
    }
    \subfigure[CelebA: Debias Teacher]{
        \includegraphics[width=0.23\textwidth]{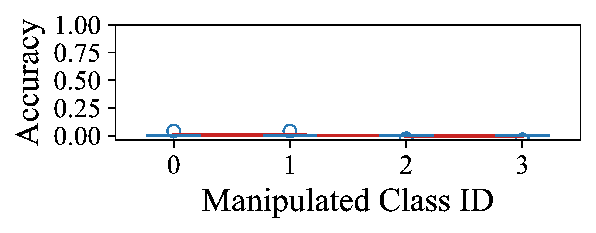}
    }
    \caption{Results of overlapping validation where we respectively manipulate each class by fake labels and report the accuracy of the target class in the testing set (the lower the better).
    We report the box plots (blue) and the average results (red) of 10 trials.
    The first, second, and third rows show the results on MNIST, CIFAR10, and CelebA respectively.}
    \label{fig:overlap}
\end{figure*}

\begin{table*}[t]
\centering
\caption{Non-target data accuracy (\%) of different types of fake labels when manipulating each class.}
\resizebox{\linewidth}{!}{
\begin{tabular}{c|cccc|cccc|cccc}  
\toprule
\multirow{2}{*}{Class ID} & \multicolumn{4}{c|}{MNIST} & \multicolumn{4}{c|}{CIFAR10} & \multicolumn{4}{c}{CelebA} \\
    & Uniform & Random & Teacher & Debias  & Uniform & Random & Teacher & Debias & Uniform & Random & Teacher & Debias \\
\midrule
0 & 98.9$\pm$0.1 & 98.9$\pm$0.1 & 98.9$\pm$0.1 & 98.9$\pm$0.1 & 91.1$\pm$0.6 & 90.8$\pm$0.6 & 90.6$\pm$0.6 & 90.8$\pm$0.7 & 95.0$\pm$0.6 & 94.9$\pm$0.5 & 95.1$\pm$0.6 & 95.2$\pm$0.7\\
1 & 98.7$\pm$0.1 & 98.7$\pm$0.2 & 98.7$\pm$0.2 & 98.8$\pm$0.1 & 89.6$\pm$0.9 & 89.5$\pm$0.7 & 89.7$\pm$0.5 & 90.0$\pm$0.6 & 95.1$\pm$0.5 & 95.0$\pm$0.6 & 95.2$\pm$0.4 & 95.2$\pm$0.5\\
2 & 99.0$\pm$0.1 & 99.0$\pm$0.2 & 98.9$\pm$0.2 & 99.0$\pm$0.1 & 92.1$\pm$0.6 & 91.9$\pm$1.0 & 91.7$\pm$0.6 & 91.6$\pm$0.9 & 95.0$\pm$0.4 & 95.2$\pm$0.7 & 94.9$\pm$0.5 & 95.1$\pm$0.7\\
3 & 98.9$\pm$0.1 & 99.0$\pm$0.1 & 98.9$\pm$0.2 & 98.9$\pm$0.1 & 93.7$\pm$0.9 & 93.7$\pm$0.8 & 91.0$\pm$0.6 & 91.8$\pm$0.6 & 95.2$\pm$0.4 & 94.8$\pm$0.4 & 95.1$\pm$0.6 & 95.2$\pm$0.6\\
4 & 98.9$\pm$0.3 & 99.0$\pm$0.1 & 98.9$\pm$0.1 & 99.0$\pm$0.1 & 90.4$\pm$1.9 & 91.2$\pm$0.8 & 91.1$\pm$0.6 & 90.4$\pm$1.2 & - & - & - & - \\
5 & 99.0$\pm$0.1 & 99.0$\pm$0.1 & 99.0$\pm$0.1 & 99.0$\pm$0.1 & 92.5$\pm$0.3 & 92.9$\pm$0.7 & 91.8$\pm$1.8 & 91.9$\pm$0.5 & - & - & - & - \\
6 & 99.0$\pm$0.2 & 99.0$\pm$0.1 & 99.0$\pm$0.1 & 98.8$\pm$0.2 & 89.8$\pm$1.7 & 90.2$\pm$1.0 & 90.4$\pm$0.5 & 90.4$\pm$0.3 & - & - & - & - \\
7 & 99.0$\pm$0.2 & 99.1$\pm$0.1 & 99.0$\pm$0.1 & 99.0$\pm$0.2 & 89.9$\pm$1.0 & 90.2$\pm$0.3 & 90.1$\pm$0.5 & 90.2$\pm$1.0 & - & - & - & - \\
8 & 99.1$\pm$0.1 & 99.1$\pm$0.1 & 99.1$\pm$0.1 & 99.0$\pm$0.1 & 89.7$\pm$0.5 & 90.5$\pm$0.5 & 89.7$\pm$0.6 & 89.8$\pm$1.0 & - & - & - & - \\
9 & 99.1$\pm$0.1 & 99.1$\pm$0.1 & 99.2$\pm$0.1 & 99.1$\pm$0.1 & 90.7$\pm$0.4 & 90.6$\pm$0.4 & 90.4$\pm$0.5 & 90.3$\pm$0.4 & - & - & - & - \\
\bottomrule
\end{tabular}
}
\label{tab:over_non}
\end{table*}

\subsubsection{Unlearning Efficiency}\label{sec:eff}
We use running time to measure the unlearning efficiency of the compared methods.
Specifically, we report the average running time of unlearning the first class in a specific client in Table~\ref{tab:time}.
As shown in Table~\ref{tab:time}, SGA, CDP, and \model{} on average improve the efficiency of Retrain by 63.73, 9.35, and 39.51 times.
On the contrary, FRR spends a significantly greater amount of time than Retrain due to the precise computation required by the Newton optimizer.
This is totally unacceptable because the design ethos behind unlearning methods is to achieve more efficient unlearning than Retrain.
Although SGA demonstrates higher efficiency compared to our proposed \model{}, it falls significantly short in terms of model utility, as we will illustrate in Section~\ref{sec:utility}.

\subsubsection{Model Utility}\label{sec:utility}
As we mentioned in Section~\ref{sec:pri}, preserving the model utility is also an important principle of federated unlearning.
An adequate unlearning method is supposed to avoid over-removing the influence of target data.
Therefore, we compare the after-unlearning model utility by evaluating the model performance on the testing set, and report the results in Table~\ref{tab:complete}.
From it, we observe that most compared methods, i.e., Retrain, FRR, and \model{}, achieve Test Acc that is close to what was achieved before unlearning.
Specifically, regarding Retrain as a baseline, FRR and \model{} can limit the difference between their Test Acc and that of Retrain to below 3.16\% and 2.32\% respectively, indicating that there is no significant forgetting of the non-target data.
However, SGA and CDP show a decline in Test Acc by 29.63\% and 10.10\% respectively, indicating a significant negative impact on model utility.

\subsection{Overlapping Validation}\label{sec:overlap}

We conduct an overlapping validation experiment for two purposes:
(i) to empirically validate the existence of overlapping solution space for tasks $\mathcal{D}'$ and $\mathcal{M}$. As we mentioned in Section~\ref{sec:know}, the existence of overlapping solution space is one of the most fundamental requirements for EWC training,
and (ii) to compare the performance of the aforementioned labels, including two benchmark labels, i.e., uniform label and random label, and two of our proposed labels, i.e., teacher label and debias teacher label.
Thus, we train the model on the mixed task $\mathcal{D}' + \mathcal{M}$, and observe the performance, which indicates the upper bound of EWC training.
For task $\mathcal{D}'$, we remove the target class $\mathcal{R}$ from the original local dataset $\mathcal{D}_k$.
For task $\mathcal{M}$, we replace the label of the target class with a fake label to construct manipulated data.
We mix the data from the above two tasks for training and test the model on the testing set.
Manipulating one class at each time, we report the accuracy of the target class in Figure~\ref{fig:overlap} and the accuracy of non-target data in Table~\ref{tab:over_non}.

If the overlapping solution space exists, the model is supposed to achieve desirable performance on both tasks, i.e., (i) on task $\mathcal{D}'$: having high accuracy on non-target data, and (ii) on task $\mathcal{M}$: having low accuracy on target class, which means it has no knowledge about target class.
From Table~\ref{tab:over_non}, we observe that all types of labels achieve high accuracy on non-target data, indicating that using whichever label can learn the knowledge from task $\mathcal{D}'$. 
From Figure~\ref{fig:overlap}, we observe that uniform and random labels fail to yield an overlapping solution space.
They suffer from the phenomenon of intransigence.
They achieve considerably high accuracy and only have low accuracy in few classes, which means they have residual knowledge of the target class. 
Their performance is also unstable, which proves that they are hard to learn.
Teacher label has low accuracy in most classes.
However, it fails in a few classes. 
This is because the untrained model potentially has prediction bias on particular classes, making the accuracy on these classes relatively high~\citep{espin2021explaining}.
Debias teacher label has low testing accuracy for all classes, proving that it exists an overlapping solution space robustly.

\subsection{Ablation Study}

\subsubsection{Memory Generator}\label{sec:mg}
The overlapping validation experiment in Section~\ref{sec:overlap} compares our proposed debias teacher label with random, uniform, and teacher labels w.r.t. the performance on the target data.
In general, the results demonstrate that the debias teacher label produces easy-to-learn labels, which facilitates the process of active forgetting.
Based on the design principle, all four types of labels are knowledge-free. However, as shown in Figure~\ref{fig:overlap}, they are not all easy-to-learn. Firstly, the comparison between purely artificial labels (uniform label and random label) and teacher-based labels (teacher label and debias teacher label) shows that purely artificial labels have worse and unstable performance while teacher-based labels achieve robust performance, indicating that the teacher-student learning pattern can generate easy-to-learn memories.
Secondly, the comparison between teacher label and debias teacher label shows that importing propensity score to debias the target class effectively addresses the issue of model bias.
 
\subsubsection{Knowledge Preserver}
To investigate the effectiveness of EWC, we compare \model{} with \model{}-C which continually trains the model with conventional loss.
As shown in Table~\ref{tab:complete} (after unlearning), \model{}-C achieves comparable DB Acc as Retrain.
But it also shows a decrease in Test Acc by 62.38\% when compared with Retrain.
This indicates that \model{}-C unlearns the target data, but also forgets some of the non-target data, which is a topical phenomenon of catastrophic forgetting.
As for comparison, our proposed \model{} utilizes EWC training to alleviate catastrophic forgetting, limiting the impact of unlearning on the target data.

\subsection{Parameter Sensitivity}\label{sec:para}

To properly achieve unlearning, our proposed framework is supposed to have (i) low accuracy on target data (ideally 0\% accuracy), and (ii) relatively high accuracy on non-target data (comparable results with the learning process).
Therefore, we choose adequate hyper-parameters, i.e., EWC training epoch and $\lambda$, based on the above two metrics.

\begin{figure}[t]
    \centering
    \subfigure[MNIST: Target Data]{\label{fig:epoch_1}
        \includegraphics[width=0.22\textwidth]{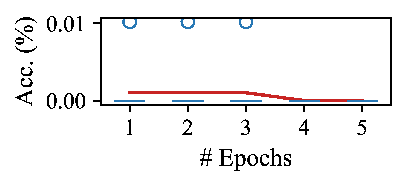}
    }
    \subfigure[MNIST: Non-target Data]{\label{fig:epoch_2}
        \includegraphics[width=0.22\textwidth]{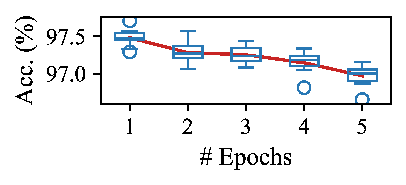}
    }
    \subfigure[CIFAR10: Target Data]{\label{fig:epoch_3}
        \includegraphics[width=0.22\textwidth]{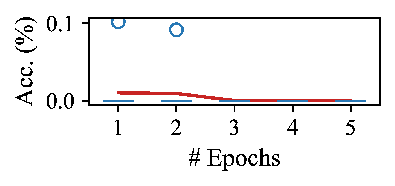}
    }
    \subfigure[CIFAR10: Non-target Data]{\label{fig:epoch_4}
        \includegraphics[width=0.22\textwidth]{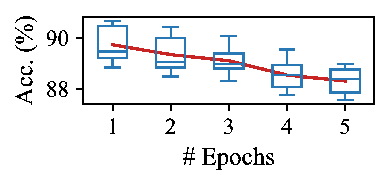}
    }
    \subfigure[CelebA: Target Data]{\label{fig:epoch_5}
        \includegraphics[width=0.22\textwidth]{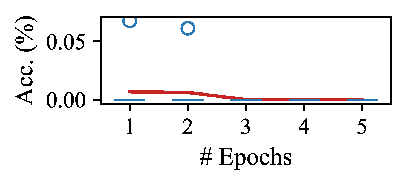}
    }
    \subfigure[CelebA: Non-target Data]{\label{fig:epoch_6}
        \includegraphics[width=0.22\textwidth]{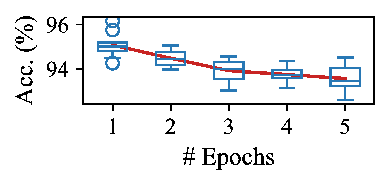}
    }
    \caption{Accuracy with different EWC training epochs.
    We report box plots (blue) and average results (red) with 10 trials.}
    \label{fig:epoch}
\end{figure}

\nosection{EWC Training Epoch} Based on the following observations in Figure~\ref{fig:epoch}, we choose the minimal epoch options, setting EWC training epoch as 1.
\begin{itemize}
    \item \textit{Target data}: As shown in Figures~\ref{fig:epoch_1}, \ref{fig:epoch_3}, and \ref{fig:epoch_5}, EWC training can already achieve almost 0\% accuracy with the minimal epoch options. 
    The accuracy is decreased to absolute 0\% when the epoch is greater than 3.
    \item \textit{Non-target data}: As shown in Figures~\ref{fig:epoch_2}, \ref{fig:epoch_4}, and \ref{fig:epoch_6}, EWC training achieves the best non-target accuracy with the minimal epoch options 
    and then gradually declines.
\end{itemize}

\begin{figure}[t]
    \centering
    \subfigure[MNIST: Target Data]{\label{fig:lam1}
        \includegraphics[width=0.22\textwidth]{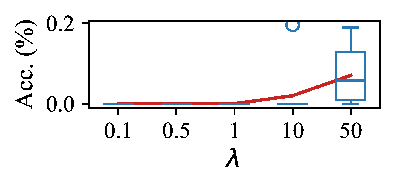}
    }
    \subfigure[MNIST: Non-target Data]{\label{fig:lam2}
        \includegraphics[width=0.22\textwidth]{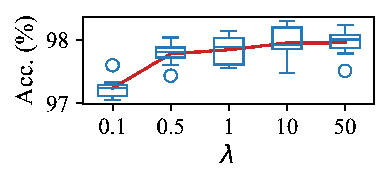}
    }
    \subfigure[CIFAR10: Target Data]{\label{fig:lam3}
        \includegraphics[width=0.22\textwidth]{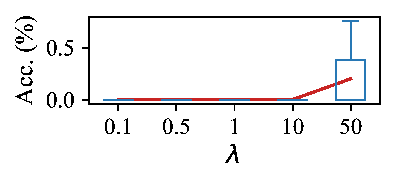}
    }
    \subfigure[CIFAR10: Non-target Data]{\label{fig:lam4}
        \includegraphics[width=0.22\textwidth]{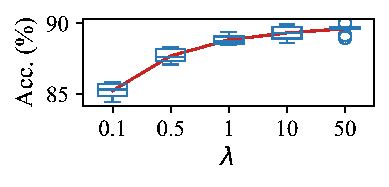}
    }
    \subfigure[CelebA: Target Data]{\label{fig:lam5}
        \includegraphics[width=0.22\textwidth]{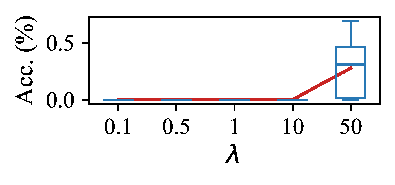}
    }
    \subfigure[CelebA: Non-target Data]{\label{fig:lam6}
        \includegraphics[width=0.22\textwidth]{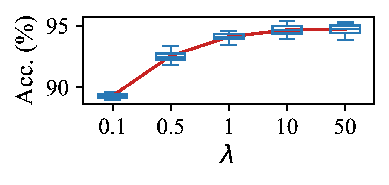}
    }
    \caption{Accuracy of with different $\lambda$.
    We report box plots (blue) and average results (red) with 10 trials.}
    \label{fig:lam}
\end{figure}

\nosection{Trade-off Coefficient} $\lambda$ balances the trade-off between unlearning target data (low accuracy on target data) and not forgetting non-target data (high accuracy on non-target data).
Based on the following observations in Figure~\ref{fig:lam}, we set $\lambda$ as 10, since it strikes a good balance.
\begin{itemize}
    \item \textit{Target data}: As shown in Figures~\ref{fig:lam1}, \ref{fig:lam3}, and \ref{fig:lam5}, when $\lambda \leq 10$, our proposed framework can achieve almost 0\% accuracy of target data. 
    The target data accuracy gradually grows when further increasing $\lambda$, which means the model does not completely unlearn the target data.
    \item \textit{Non-target data}: As shown in Figures~\ref{fig:lam2}, \ref{fig:lam4}, and \ref{fig:lam6}, the accuracy of non-target data increases with the value of $\lambda$, but the degree of increase decreases gradually.
\end{itemize}

\subsection{Unlearning Multiple Classes}

In real-world scenarios, unlearning requests are often submitted sequentially, making batch processing impractical. To address this relatively overlooked yet practically significant issue, we conducted experiments on sequential unlearning of multiple classes. As shown in Table~\ref{tab:complete_mul}, we sequentially unlearned 3 classes to evaluate the effectiveness of our approach.
From the results, we observe trends consistent with those in Table~\ref{tab:complete} (i.e., unlearning a single class). Specifically, our proposed FedAF achieves comparable BD Acc to Retrain while maintaining high Test Acc. This observation indicates that FedAF effectively achieves satisfactory unlearning completeness without significantly compromising model utility. These findings further validate the robustness and practicality of our proposed method in handling sequential unlearning scenarios, demonstrating its potential for real-world applications where unlearning requests are frequent and diverse.

\begin{table*}[t]
\centering
\caption{Accuracy (Acc) of unlearning multiple classes sequentially.
The BD Acc (the lower the better) and the Test Acc are used to evaluate unlearning completeness and model utility, respectively.
We \textbf{bold} the top results of each class, other than Retrain.}
\resizebox{\textwidth}{!}{
\begin{tabular}{c|c|cccccc|cccccc}  
\toprule
\multirow{2}{*}{Class ID} & \multicolumn{1}{c|}{BD Acc (\%)} & \multicolumn{6}{c}{BD Acc (\%) after unlearning $\downarrow$} & \multicolumn{6}{c}{Test Acc (\%) after unlearning $\uparrow$}\\
    & before unlearning & Retrain & FRR & SGA & CDP & \model{}-C & \model{} & Retrain & FRR & SGA & CDP & \model{}-C & \model{}\\
\midrule
\multicolumn{14}{c}{MNIST}\\
\midrule
0       & 81.14$\pm$2.86 & 0.12$\pm$0.01 & 0.73$\pm$0.13 & 2.71$\pm$0.94 & 65.43$\pm$1.12 & 0.60$\pm$0.35 & \textbf{0.58$\pm$0.18} & 96.83$\pm$0.13 & 91.14$\pm$0.12 & 71.32$\pm$2.21 & 84.39$\pm$1.13 & 40.03$\pm$0.51 & \textbf{92.22$\pm$0.26}\\
0, 1    & 81.35$\pm$3.14 & \textbf{0.12$\pm$0.14} & 0.76$\pm$0.46 & 2.57$\pm$0.51 & 68.93$\pm$1.32 & 0.67$\pm$0.51 & 0.78$\pm$0.41 & 96.83$\pm$0.14 & 91.01$\pm$0.14 & 70.43$\pm$1.63 & 84.02$\pm$1.35 & 40.00$\pm$0.51 & \textbf{92.25$\pm$0.43}\\
0, 1, 2 & 81.43$\pm$3.19 & \textbf{0.11$\pm$0.22} & 0.92$\pm$0.32 & 2.15$\pm$1.05 & 70.03$\pm$1.54 & 0.62$\pm$0.23 & 0.72$\pm$0.32 & 96.80$\pm$0.14 & 90.89$\pm$0.17 & 68.65$\pm$1.67 & 84.12$\pm$1.51 & 39.85$\pm$0.41 & \textbf{92.14$\pm$0.31}\\
\midrule
\multicolumn{14}{c}{CIFAR10}\\
\midrule
0       & 71.00$\pm$1.73 & 0.01$\pm$0.00 & 1.64$\pm$1.05 & \textbf{0.00$\pm$0.00} & 56.35$\pm$1.33 & 1.45$\pm$1.87 & 1.45$\pm$1.02 & 81.03$\pm$0.63 & \textbf{80.04$\pm$0.61} & 61.82$\pm$1.19 & 72.46$\pm$1.06 & 29.91$\pm$6.74 & 79.55$\pm$1.92\\
0, 1    & 70.80$\pm$3.22 & 0.92$\pm$1.01 & 2.12$\pm$1.87 & \textbf{0.00$\pm$0.00} & 55.75$\pm$1.92 & 2.01$\pm$1.67 & 2.42$\pm$1.52 & 80.92$\pm$0.61 & \textbf{80.05$\pm$0.45} & 62.62$\pm$1.46 & 72.62$\pm$0.84 & 28.83$\pm$5.56 & 79.62$\pm$0.56\\
0, 1, 2 & 71.92$\pm$3.53 & 0.89$\pm$1.14 & 2.56$\pm$1.25 & \textbf{0.00$\pm$0.00} & 55.34$\pm$2.27 & 1.92$\pm$1.59 & 2.36$\pm$1.41 & 81.05$\pm$0.87 & 80.10$\pm$0.65 & 65.62$\pm$1.53 & 72.46$\pm$0.83 & 28.45$\pm$4.15 & \textbf{80.15$\pm$0.63}\\
\midrule
\multicolumn{14}{c}{CelebA}\\
\midrule
0       & 74.16$\pm$0.37 & 0.68$\pm$0.30 & 2.18$\pm$1.30 & \textbf{0.00$\pm$0.00} & 64.56$\pm$0.27 & 1.76$\pm$1.31 & 1.91$\pm$1.32 & 88.71$\pm$0.36 & \textbf{88.69$\pm$0.39} & 32.07$\pm$3.85 & 87.18$\pm$0.22 & 31.29$\pm$4.91 & 88.44$\pm$0.47\\
0, 1    & 74.37$\pm$1.14 & 0.63$\pm$0.43 & 2.36$\pm$1.25 & \textbf{0.00$\pm$0.00} & 60.72$\pm$0.72 & 1.72$\pm$1.15 & 1.82$\pm$1.16 & 89.15$\pm$0.41 & 88.85$\pm$0.43 & 32.53$\pm$2.51 & 87.36$\pm$1.23 & 32.01$\pm$3.14 & \textbf{89.02$\pm$0.97}\\
0, 1, 2 & 77.76$\pm$2.52 & 1.17$\pm$0.64 & 2.43$\pm$1.36 & \textbf{0.00$\pm$0.00} & 58.42$\pm$1.15 & 2.51$\pm$2.12 & 2.14$\pm$1.46 & 87.16$\pm$0.26 & \textbf{87.56$\pm$0.19} & 33.23$\pm$2.23 & 87.27$\pm$1.25 & 31.43$\pm$3.43 & 87.27$\pm$0.75\\
\bottomrule
\end{tabular}
}
\label{tab:complete_mul}
\end{table*}

\section{Conclusion}
In this paper, we propose \model{}, a novel federated unlearning method inspired by active forgetting in neurology. 
This approach is designed to be scalable for fine-grained unlearning targets and is not dependent on specific model architectures.
Generally, the core idea of active forgetting is leveraging new memories to overwrite the old ones.
This overcomes the limitations of existing approaches without storing historical updates or estimating data influence. 
Specifically, \model{} builds up on incremental learning and consists of two modules per client, i.e., memory generator and knowledge preserver. 
The memory generator produces fake labels using teacher-student learning, paired with target data features to create new memories. 
The knowledge preserver trains the model with these new memories using refined EWC training to mitigate catastrophic forgetting. 
Experimental results on three benchmark datasets show that \model{} efficiently unlearns target data of specific class within the client from the global model while preserving non-target data knowledge.
A comprehensive ablation study further validates the crucial role of the memory generator and knowledge preserver in achieving effective unlearning, providing insights into the contributions of each component to the overall performance of \model{}.
Additionally, we conduct backdoor attacks to evaluate the completeness of unlearning, and the results demonstrate the effectiveness of \model{}.

\section{Limitation and Broader Impact}
While our proposed neuro-inspired federated unlearning framework demonstrates promising results in achieving fine-grained unlearning targets and mitigating catastrophic forgetting, several practical issues remain to be addressed. For instance, in real-world scenarios, the data features of the target class to be unlearned may not be available during the unlearning process, resulting in a data-free unlearning task. However, our proposed framework relies on the original data features to construct new memories, which limits its direct applicability in such scenarios. In our future work, we plan to explore leveraging class features in the embedding space to circumvent the need for original data features, thereby enhancing the framework's versatility.
\lyy{Additionally, the scalability of the proposed framework to large-scale federated networks remains a challenge. As the number of participating clients grows, the impact of local client updates on the global model diminishes, potentially leading to slower convergence of unlearning. In future work, we plan to investigate these practical issues, e.g., managing large numbers of clients and dynamic client participation.}

The active forgetting approach not only achieves fine-grained unlearning in federated learning but also provides valuable insights for general unlearning tasks and other specific unlearning scenarios. By addressing these limitations, we aim to further advance the practicality and applicability of unlearning techniques in real-world applications.

\section*{Declaration of Interests}
The authors declare that they have no known competing financial interests or personal relationships that could have appeared to influence the work reported in this paper.



\end{document}